\documentclass[twoside,11pt]{article}

\usepackage{jmlr2e}
\usepackage[utf8]{inputenc}

\usepackage{amsmath}
\usepackage{amssymb}
\usepackage{stackrel}
\usepackage[colorinlistoftodos]{todonotes}

\ShortHeadings{Robustness Predictions Natural Perturbations}{Scher and Tr\"ugler}
\firstpageno{1}

\begin{document}

\title{Testing robustness of predictions of trained classifiers against naturally occurring perturbations}

\author{\name Sebastian Scher \email sscher@know-center.at \\
       \addr Know-Center GmbH\\
       Inffeldgasse 13/6\\
       8010 Graz, Austria
       \AND
       \name Andreas Tr\"ugler \email atruegler@know-center.at \\
       \addr Know-Center GmbH\\
       Inffeldgasse 13/6\\
       8010 Graz, Austria\\
       and\\
       Institute of Interactive Systems and Data Science\\
	   Graz University of Technology\\
	   Inffeldgasse 16C, 8010 Graz, Austria\\
       and\\
       Department of Geography and Regional Science\\
	   University of Graz\\
	   Heinrichstraße 36, 8010 Graz, Austria     
       }

\maketitle

\begin{abstract}

Correctly quantifying the robustness of machine learning models is a central aspect in judging their suitability for specific tasks, and ultimately, for generating trust in them. We address the problem of finding the robustness of individual predictions. We show both theoretically and with empirical examples that a method based on counterfactuals that was previously proposed for this is insufficient, as it is not a valid metric for determining the robustness against perturbations that occur ``naturally'',  outside specific adversarial attack scenarios. We propose a flexible approach that models possible perturbations
in input data individually for each application. This is then combined
with a probabilistic approach that computes the likelihood that a
``real-world'' perturbation will change a prediction, thus giving quantitative information of the robustness of individual predictions of the trained machine learning model. The method does not require access to the internals of the classifier and thus in principle works for any black-box model. It is, however, based on Monte-Carlo sampling and thus only suited for input spaces with
small dimensions. We illustrate our approach on the Iris and the Ionosphere datasets, on an application predicting fog at an airport, and on analytically solvable cases.
\end{abstract}


\section{Introduction}
``Robustness'' is a term widely used in the context of machine-learning
(ML) and statistics. In ML, it is used for several connected but at the same time distinct concepts. One usage is to refer how robust a model-prediction
is to specific perturbations in the input data. A widely studied field
is \emph{adversarial robustness}, closely related to adversarial examples
(e.g. \citealp{biggio2013evasion,biggio2013security,szegedyIntriguingPropertiesNeural2014a,goodfellowExplainingHarnessingAdversarial2015a}; building on earlier work by \cite{dalvi2004adversarial, lowd2005adversarial} also see \cite{DBLP:journals/corr/abs-1712-03141} for an overview).
Adversarial robustness (first described in \cite{szegedyIntriguingPropertiesNeural2014a}
for neural networks) deals with the susceptibility of an ML model
to perturbations that were specifically crafted to ``fool'' the
model --- so called adversarial attacks. Here the threat-model is that
an attacker starts from a correctly classified input (e.g. an image),
and then modifies the input image with a minimal --- often imperceptible -
perturbation that still leads to a misclassification. These perturbations
typically look random, but they are in fact not random and exploit
specific characteristics of the ML model and/or the training data. Adversarial robustness can be computed over a whole test set --- thus answering how robust a classifier is in general. It can also be used to compute the robustness of single test samples. The latter has been suggested as a generic robustness metric \citep{sharmaCERTIFAICommonFramework2020a}.
Another approach to analyse the robustness of classifiers is to investigate how robust the classifier is to Gaussian noise in the input data, whereas the noise is modelled in a way that resembles naturally occurring distortions (e.g. \citep{rusak2020simple}). This allows to compute the robustness of a classifier in general and to compare the robustness of different classifiers.
Previous work has either focused on how to compute robustness of classifiers in general --- including robustness to naturally occurring perturbations --- or on computing the robustness of individual predictions, the only using methods for adversarial robustness. 

In this paper, we deal with a related but different question: How can we assess the robustness
of individual predictions of an ML-model to perturbations that are not crafted by an adversarial
attacker, but that occur by chance and/or accident? We will call these
perturbations \emph{real-world-perturbations}. Such perturbations
could occur --- among others --- from: noise, measurement errors, data-drift
and data-processing errors. In this paper we will show why adversarial
robustness is not necessary a valid measure for the robustness against
such perturbations. Throughout this paper, the term \emph{real-world-perturbations} will only refer to this type of perturbations.

It is impossible to completely separate our idea of \emph{real-world-perturbations} 
from other central aspects in ML, namely generalization and distribution-shift.
Additionally, the question of how to acquire a suitable test-set is
related, as well as the concept of counterfactuals and the ideas behind
data-augmentation (including test-time augmentation \citep{shanmugamWhenWhyTestTime2020a}).

\subsection{Contributions} 
We start our discussion with a different angle of view compared to earlier works and focus on the main question: 
\begin{center}
Given a trained ML application and a test-dataset, how can we assess the robustness of \emph{individual predictions}
to perturbations that could occur when deploying the model in a real
application, but are not represented in the test set?
\par\end{center}
These perturbations need not be from a fixed distributions (e.g. noise with a certain variance), but can vary from point to point, and from setting to setting.

To answer this question, in this paper we will
\begin{itemize}
\item define \emph{real-world-robustness} in a mathematical way, based on previous work,
\item show why adversarial robustness and metrics based on counterfactuals
are insufficient for determining real-world-robustness,
\item show how to compute real-world-robustness for low-dimensional datasets,
\item compare adversarial robustness with real-world-robustness in three examples,
\item discuss issues surrounding transferring the ideas to high-dimensional
problems.
\end{itemize}

Another major difference compared to much of the previous work on the topic is that our approach is independent (or in other words, completely ignorant) of the training data, and has thus to be differentiated from all related concepts that have something to do with training data (e.g. robustness to noise in training data). Our method starts from a trained classifier, where this classifier comes from does not matter. It could even be a "hand-coded" one (e.g. symbolic AI). Noise tolerance in this sense is for example also an issue of lifelong learning machines \citep{Kudithipudi2022}.
Our method also differs from previously proposed methods by focusing on whether the prediction changes or not, independent of whether the base prediction is correct or not, and by not concerning on how the accuracy of a prediction changes. There are settings in which this question is highly relevant, especially in applications and cases were it would be hard to tell whether a prediction of a model is correct or not. One example is an insurance that uses a classification model to assess whether a potential customer is eligible for a certain insurance product. If a client is rejected by the model, and challenges this decision, then it would be beneficial for the insurance company to be able to show that the decision would not have changed, even if the input parameters would have been slightly different. Thus, as a pre-caution, the insurance company might want to check every single prediction on its robustness, and potentially deal with the less robust ones in a different way (e.g. make an additional assessment by a human), in order to avoid liability problems.

This work is addressed both to the research community and to practitioners
in the growing field of testing and certifying AI-applications \citep{supremeauditinstitutionsoffinlandgermanythenetherlandsnorwayandtheukAuditingMachineLearning2020a, winterTrustedArtificialIntelligence2021a}. Therefore, this paper also
aims to convey an intuitive understanding of the issues surrounding
robustness of ML models.

\subsection{Taxonomy}
We use the term robustness in the above described meaning. It should be noted that sometimes \emph{robustness} refers to how well a model works in other settings, and then the word \emph{reliability} is used instead to describe the sensitivity to small changes in the input (e.g. in the NIST draft for a taxonomy of AI risk \citep{nistDraftTaxonomyAI2021a}).

\section{Related Concepts and Related Work}

In this section we discuss concepts that are related to our problem
in the existing literature. A recent review \citep{huang2020} also gives a broad overview on safety and trustworthiness of deep neural networks with focus on related topics like verification, testing, adversarial attack and defence, and interpretability.

\subsection{Perturbation Robustness}
The term perturbation robustness was introduced by \cite{hendrycks2019benchmarking}, and used in subsequent work \cite{rusak2020simple}. It is the basis for our work. The authors defined the term as the ``flip-probability'' of a network under perturbations of the inputs. They compute this flip-probability over the whole test set, and with this aggregated metric show that neural networks for image recognition tasks that are robust to adversarial examples are not necessarily robust against naturally occurring perturbations. While they show that for a neural network classifier in general adversarial robustness and perturbation robustness are not the same, they do not deal with the question whether the robustness of individual predictions, including the ranking of predictions by robustness, is comparable when using adversarial robustness and perturbation robustness, or not. This is what we do in our paper. Our work starts from their definition of perturbation robustness, which we reformulate so that it is more intuitive for our task at hand (see Section 3.2).

\subsection{Common Corruptions}
The term "common corruptions" was introduced by \cite{hendrycks2019benchmarking}, and is very similar to their concept of perturbation robustness. In their work, common corruptions are "typical" corruptions than can occur on images. They are not modelled on a sample-by-sample basis, but are used for all test samples. As for perturbation robustness, they compute the classification error over the whole test set. In contrast to \cite{hendrycks2019benchmarking}, in our work, we deal with perturbations that are potentially different for each test sample.

\subsection{Adversarial Examples, Counterfactuals and Adversarial Robustness}

Adversarial examples are samples of input for a ML classification system that are very similar to a normal input example, but cause the ML system to make a different classification. A typical example are images that are slightly modified (often so slightly that the difference is invisible by eye) and then lead to a misclassification. The concept is generic and not limited to image recognition systems. Such adversarial examples exploit certain properties of ML classifiers, and are explicitly and purposefully searched with specific algorithms (called adversarial attacks). There is a lot of discussion in the literature on why such examples do exist at all.
In \cite{szegedyIntriguingPropertiesNeural2014a} the authors suggested that adversarial examples could be of extremely low probability. If this is true, then
one would not expect that they occur by chance (i.e. outside an adversarial
attack scenario). \cite{goodfellowExplainingHarnessingAdversarial2015a},
however, disputed this low-probability explanation of adversarial
examples, and instead argued that they are caused by the local linearity
of the neural networks. Other perspectives/explanations have been offered as well \citep{ilyasAdversarialExamplesAre2019,tanayBoundaryTiltingPersepective2016}.
There is also literature on ``real-world adversarial attacks'' that deals with adversarial attacks in real-world scenarios \citep{kurakinAdversarialExamplesPhysical2017a,eykholtRobustPhysicalWorldAttacks2018a}. This has, however, nothing to do with our definition of real-world-robustness, as it still considers the adversarial threat model.

Counterfactuals (also known as counterfactual explanations~\citep{Kment2006-KMECAE}) are very similar to adversarial examples. The idea is to find, for a given test data point, a different data point with a different classification (either with a different classification of a specified class, or with any classification different than the classification of the test point), under the constraint that this example should be as similar as possible as the original test point. It can thus be used for explaining classifications in a certain sense (``if feature x had a value of b instead of a, then the classification would be different''). Such examples are called counterfactuals. They have been first proposed in order to fullfill the ``right to an explanation'' in the General Data Protection Regulation (GDPR) \citep{wachterCounterfactualExplanationsOpening2018a, 10.1145/3287560.3287574}. 

Adversarial examples and counterfactual explanations are highly related,
up to the point that they are sometimes seen as the same thing with
two different names. Undisputed is that both stem from the same mathematical
optimization problem, and are thus mathematically the same thing.
Whether they are also \emph{semantically} the same is part of ongoing
discussions as well. See \cite{freieslebenIntriguingRelationCounterfactual2021a,katzReluplexEfficientSMT2017}
for a thorough discussion on the topic. Here, we do not attempt to
answer that question, but instead use the terms interchangeably, as for our setting the (potential) distinction is not important.

Usually, counterfactuals (and adversarial examples) require a pre-defined distance metric. However,
\cite{pawelczykLearningModelAgnosticCounterfactual2020a} proposed
a method that \emph{learns} a suitable distance metric from data. Many different methods for generating counterfactuals and adversarial examples are known.  Ten 
alone are integrated into the CARLA python library \citep{pawelczykCARLAPythonLibrary2021a}, and additional ones in the Adversarial Robustness Toolbox \citep{nicolaeAdversarialRobustnessToolbox2019}.
Some methods work only on specific model types, others on black box
models (like CERTIFAI \citep{sharmaCERTIFAICommonFramework2020a})

In our paper, we discuss adversarial examples and counterfactuals
agnostic of they way they are be produced --- we simply assume that
they could somehow be obtained.

Together with the idea of adversarial attacks, the concept of measuring \emph{adversarial robustness} emerged. This is a  measure to quantify how robust (or susceptible) a classifier is to adversarial attacks. Two examples of ways to estimate \emph{adversarial robustness}  are the CLEVER score, which is an 
attack-agnostic robustness metric for NNs \citep{wengEvaluatingRobustnessNeural2018a}, and the CERTIFAI-robustness score for black box models, which assigns a robustness score for individual data points based on the distance to the closest counterfactuals using a genetic algorithm \citep{sharmaCERTIFAICommonFramework2020a}.

\subsection{Perturbation Theory and Interpretability Methods}
If small perturbations are introduced to a physical system (e.g. noise, fluctuations, etc.) it is often very helpful to develop a truncated power series expansion in terms of these perturbations in order to find approximate solutions of the system. The perturbed system can then be described as a simpler (ideally mathematical directly solvable) system plus corresponding correction terms, a very common approach in quantum mechanics for example. The mathematical foundation behind is called \emph{perturbation theory} and the general question of how (small) perturbations are affecting a system of interest is obviously also linked to our discussion here. While a direct assessment of a perturbation theoretic approach is beyond the scope of our paper, at least related concepts have found their way to modern machine learning literature as well. Investigating the influence of perturbations of input data~\citep{olden2002illuminating} or analyzing machine model behaviour based on power series decomposition~\citep{MONTAVON2017211} have been introduced especially in the field of explainable AI for example. One of the most popular methods to identify the influence of specific input features in the output of a machine learning model (so-called attribution methods) is based on perturbing the input. A description of \emph{CAPTUM}, an open-source library for interpretability methods containing generic implementations of a number of gradient and perturbation-based attribution algorithms, as well as a set of evaluation metrics for these algorithms, can be found in~\cite{captum}. Such approaches usually focus on ``faithfullness'', e.g. the ability of a machine learning algorithm to recognize important features, instead of the overall robustness of the model.


\subsection{Data Augmentation}
Data augmentation are techniques to artificially create new training samples from existing training data. It is widely used in tasks that work with images, as for image recognition it is relatively easy to find modifications that change the image, but in a way that the image still contains the same semantic information. Data augmentation is used to prevent overfitting. Therefore, it is indirectly related to robustness, as the robustness of a classifier also depends on its generalization properties, but it cannot be used as a measure for robustness. Data augmentation is usually used during model training and not when the trained model is evaluated. There is, however, also the idea of test-time data augmentation \citep{shanmugamWhenWhyTestTime2020a}. Here ``test-time'' refers to making predictions with the model after training. This is used because it can help improve the predictions, but it is also only indirectly related to robustness.

\subsection{Distribution Shift}
Distribution shift~\citep{quinonero2008dataset, torralba2011unbiased, bengio2019metatransfer, federici2021informationtheoretic, Hendrycks_2021_ICCV} is the phenomenon when the training data does not completely represent the test data that is encountered in a real world application. A curated benchmark of various datasets reflecting a diverse range of naturally occurring distribution shifts (such as shifts arising from different cameras, hospitals, molecular scaffolds, experiments, demographics, countries, time periods, users, and codebases) can be found in~\cite{pmlr-v139-koh21a}. Robustness against distribution shift~\citep{NEURIPS2020_d8330f85} is closely related to the ability of a ML classifier to \emph{generalize} to points beyond the training examples and also to overfitting as a counterpart to generalization.\\
Our definition of real-world-robustness differs from distribution shifts --- the main difference is that the study of distribution shifts is primarily concerned with the accuracy of classifiers as a whole, whereas we are concerned with the robustness of individual predictions. It is not true in a general sense that models that generalize well are automatically robust within our real-world-robustness definition. Still, distribution shifts are also a relevant issue when determining real-world-robustness in our approach. We model possible real world distributions, but when the application is put in practice, it might be that the actual occurring perturbations are different --- thus representing a distribution shift in the possible perturbations.
Yet another similar topic is robustness towards out-of-distribution points \citep{yang2021generalized}, which focuses on unusual data that was not observed in the training dataset (e.g. a self-driving cat that detects a new object type it has never seen before).

\subsection{Certification of Neural Networks}

For neural networks (NNs) there is a lot of active research on finding formal guarantees and thus ``proving'' that a network is robust to particular perturbations and/or fulfills
other properties. This is often referred to as ``verification''
or ``certification'' \citep{singhAbstractDomainCertifying2019a}.
For example, it is possible to find robustness guarantees for NNs with defining hypercubes around test points, and
then test whether the classification
stays the same for all areas within this hypercube. At the moment, this is only possible with strong assumptions and restrictions on the possible perturbations (e.g. \citep{cohenCertifiedAdversarialRobustness2019b,singhAbstractDomainCertifying2019a}).
\cite{huangSafetyVerificationDeep2017b} proposed a method that can offer high guarantees, but under the restriction that input perturbations need to be discretized.

\subsection{Robustness to Noise in Training Data}

The topic of noisy data has long been studied in robust statistics (e.g. \cite{ronchetti2009robust, huber2011robust}. \cite{saezEvaluatingClassifierBehavior2016} define robustness as ``A robust
classification algorithm can build models from noisy data
which are more similar to models built from clean data.'' This robustness against noise refers to the ability of the training algorithm to deal with noisy training data (noise can be on the features and/or on the labels). 
 The Relative Loss of Accuracy (RLA) score proposed by \cite{6121827} and the Equalized Loss of Accuracy (ELA) score proposed by \cite{saez2016evaluating} are ways of measuring this type of robustness (specifically to noise on the class labels). This idea of robustness, however, is fundamentally different from our definition of robustness, which deals with robustness of the trained classifier against noise (and other perturbations) added to test samples.
 A systematic overview of the impact of noise on classifiers is given in \cite{zhu2004class}, and the effect of label noise on the complexity of classification problems in \cite{garciaEffectLabelNoise2015}.

\subsection{Domain Generalization}
This topic is closely related to distribution shifts. Domain generalization describes the ability of an algorithm to make predictions on a different domain than it was trained on. Early work on the topic was done by \cite{blanchard2011generalizing,daomaingen2013} . More recently, \cite{DBLP:journals/corr/abs-2007-01434} discussed the importance of model selection for domain generalization, and introduced a testbed for it.

\section{Methods}

In this section we formalize our question on how to asses real-world-robustness, show why adversarial samples and counterfactuals are insufficient
for that purpose, and discuss how to compute real-world-robustness.

\subsection{ Adversarial Examples and Counterfactuals}

Untargeted counterfactuals $x_c$ (or untargeted adversarial examples) with respect to a test point $x$ are defined via the following minimization problem (e.g. \cite{freieslebenIntriguingRelationCounterfactual2021a}):

\begin{equation}
    \text{minimize} \; d\left(x_c,x \right) \: \text{subject to} \: f\left(x_c\right)\neq f\left(x\right),
    \label{eq:CF}
\end{equation}
where $d$ is a distance function that needs to be pre-defined and $f$ is the corresponding machine learning model.

There is also the concept of \emph{targeted} counterfactuals or adversarial attacks, in which the goal is not only that the counterfactual has a different predicted label, but a specific different one. In this paper we only speak about untargeted counterfactuals and adversarial examples. For binary classification problems (which will be used in this study as examples), targeted and untargeted counterfactuals are per definition the same. Many different methods have been proposed to (approximately) solve this minimization problem (see references in Introduction).

\subsection{Real-world-robustness}


Following the definition of perturbation robustness from \cite{hendrycks2019augmix}, we reformulate it in a way that is more suited for our question of robustness of individual test points with potentially different uncertainty for each point.
Given a trained machine learning model $f$,
a test-data point\footnote{We omit a vector notation for simplicity, all variables with $x$ in our discussion are typically vectors with dimension of the input space
of the machine-learning model $f$.} $x_{t}$ and the accompanying predicted label
$y=f\left(x_{t}\right)$, how can we assess how robust $y$
is to perturbations $x'$ of $x_{t}$? We consider $x'$ not in an abstract way (e.g. a hypersphere), but as originating from
real-world processes. These can be, for example, noise or measurement
errors.

We will use $x'$ for the perturbation,
and $\hat{x}$ for the perturbed value, thus
\begin{equation*}
\hat{x}=x_{t}+x'.
\end{equation*}

Further, the perturbations $x'$ are not given in absolute values, but are
described probabilistically, leading to a $N$-dimensional probability
density function $\hat{p}\left(x'\right)$, which assigns a probability
density to each point $x$ in the input space of the model. The input space has dimension $N$, where $N$ is the number of input features. Additionally,
as in the real world we cannot expect that potential perturbations
are the same for all data-points, we introduce $\hat{p}$ as a function dependent on
$x_{t}$, which we denote as $\hat{p}_{x_t}$
\begin{equation*}
\hat{p}(x')\to\hat{p}_{x_t}\left(x'\right)
\end{equation*}

Thus each test-point $x_{t}$ gets assigned a probability-distribution
(defined over the input space $\mathbb{R}^{N}$).

To consider the example of measurements from temperature sensors,
the noise of the sensors might be different for high and for low temperatures.
Or, another example, there could be hard boundaries in some cases
(e.g. a percentage range cannot be lower than 0 or higher than 100).

The first challenge is thus to model $\hat{p}_{x_t}\left(x'\right)$.
In principle there are two ways to model $\hat{p}_{x_t}\left(x'\right)$:
(1) analytically or (2) as a process that generates samples (e.g.
via a Monte-Carlo simulation). Both will require expert knowledge,
and details of the specific application.

The second challenge is to find out whether our ML-model $f$ yields
different predictions in the region of interest around $x_{t}.$ Here
we intentionally use the vague term \emph{region of interest}, as defining
this is one of the main problems, and the approach we choose is part
of what distinguishes this work from previous work. One possible ansatz is to define hard boundaries for $x'$ (and subsequently for $\hat{x}$),
and then try to find out whether the prediction for $f$ is the same
for the whole region defined in this way, yielding a binary measure
(either whole region yields same prediction, or not). This is one
of the things that previous work on certifying neural networks are
aiming at (e.g. \cite{singhAbstractDomainCertifying2019a}.). This
approach would be, however, contrary to our idea of describing possible
perturbations as probability distributions. We thus formulate the
problem the following way:
\begin{center}
How likely is it that the model prediction changes under the assumed
perturbations?
\par\end{center}

In mathematical terms, we want to know the probability that $f\left(\hat{p}_{x_t}\left(x'\right)\right)$
returns a prediction label different from $f\left(x_{t}\right)$.
For this we can define a binary function that returns one if the prediction
is different than the one for the original data point, and zero otherwise:
\begin{equation}
L\left(x_{t},x'\right)=\begin{cases}
0 & f\left(x_t+x'\right)=f\left(x_{t}\right)\\
1 & f\left(x_t+x'\right)\ne f\left(x_{t}\right)
\end{cases}
\label{eq:binary}
\end{equation}
Then we can, in principle, integrate the product of 
 $L\left(\left(x_{t},x'\right)\right)$ and $\hat{p}_{x_t}\left(x'\right)$
over the whole input space and thus obtain the probability for a different prediction compared to the unperturbed data point:
\begin{equation}
P\left(f\left(\hat{x}\right)\ne f\left(x_{t}\right)\right)= \frac{\underset{\mathbb{R}^{N}}{\int}L\left(x_{t},x'\right)\,\hat{p}_{x_t}\left(x'\right)dx'}{\underset{\mathbb{R}^{N}}{\int}L\left( x_t,x' \right)dx'}
\label{eq:p-integral}
\end{equation}

In order to define a robustness measure we can compute 
\begin{equation*}
    P_r=1-P
\end{equation*}
This is what from now we will refer to as
\emph{real-world-robustness}\@.

\begin{figure}
\centering
\includegraphics[width=0.5\textwidth]{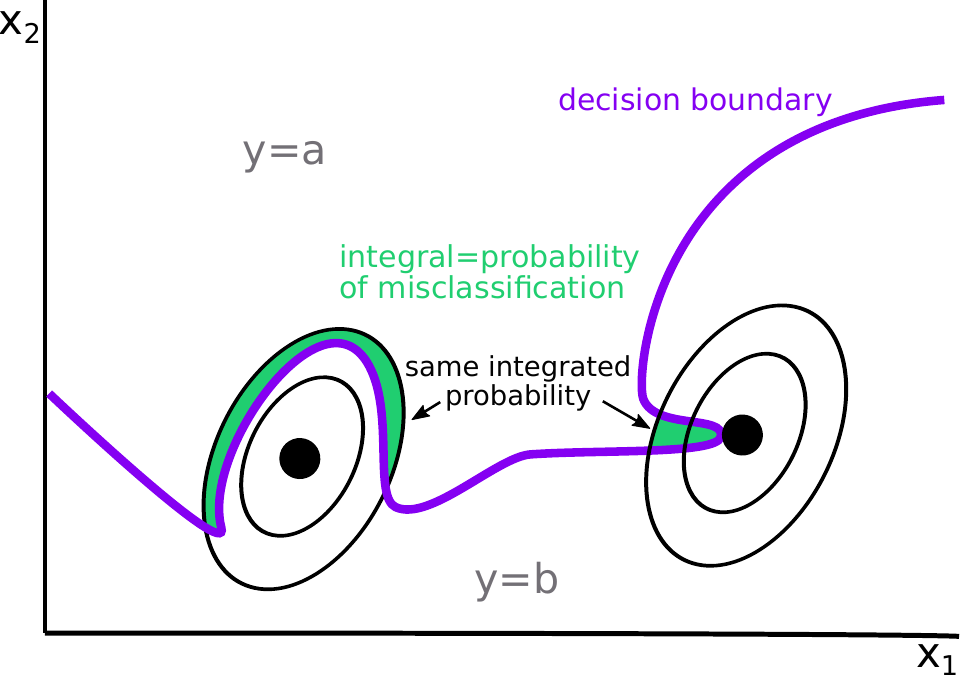}\caption{\label{fig:comparison-situations}Example of a binary classifier with two-dimensional feature space. Two datapoints (black dots) with
uncertainty around them --- here an elliptical distribution, the ellipses
are lines of equal likelihood. The purple line is the decision boundary
of a binary classifier. The left point shows a situation where all
possibilities of a misclassification are not very likely (all in regions
of the elliptical distribution with low probability), but these regions
summed together are in fact quite large, thus the probability of misclassification
is not that low (indicated by the green area). In other words, one
needs to get relatively far away from the original data point to get
a misclassification, but this can happen in many different directions.
The right data-point illustrates a different scenario: here only in
a small range of directions a misclassification is possible, but in
those directions a misclassification is happening in regions of high
likelyhood.}

\end{figure}

An important thing to realize is that there are many possible ways
that $P$ could exceed a certain value: there could be a region on the
other side of the decision boundary (thus in a region where the predicted
label is different) that is small but has high probability, but equally
well there can be situations where the density of $\hat{p}$ is relatively
small on the other side of the decision boundary, but that the region
is actually large. In the first situation, there are not many different
possible points around $x_{t}$ that could cause a wrong prediction,
but those points are not very unlikely. In the second situation, the
points that would cause a wrong prediction are all relatively unlikely,
but there are many of them, thus making the probability of getting
\emph{one} of them in a real setting is actually equally high as in
the first situation. This is illustrated in fig.~\ref{fig:comparison-situations}
for a binary classifier. For simplicity, we will show only illustrations
of classifiers with a single decision boundary --- all the principles
here also apply to more complicated settings were the decision
boundary consists of multiple unconnected pieces.

\subsection{Why Counterfactuals are not Enough}
\label{sec:Why-counterfactuals-are-not-enough}

\begin{figure}
a)\includegraphics[width=0.46\textwidth]{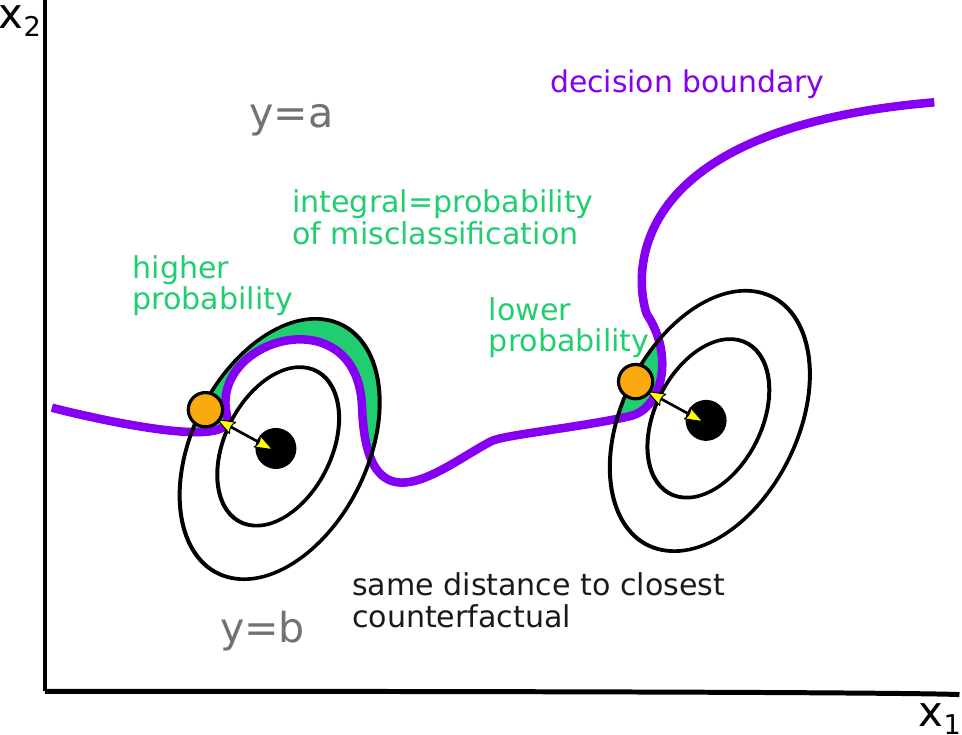} $\;$b)\includegraphics[width=0.46\textwidth]{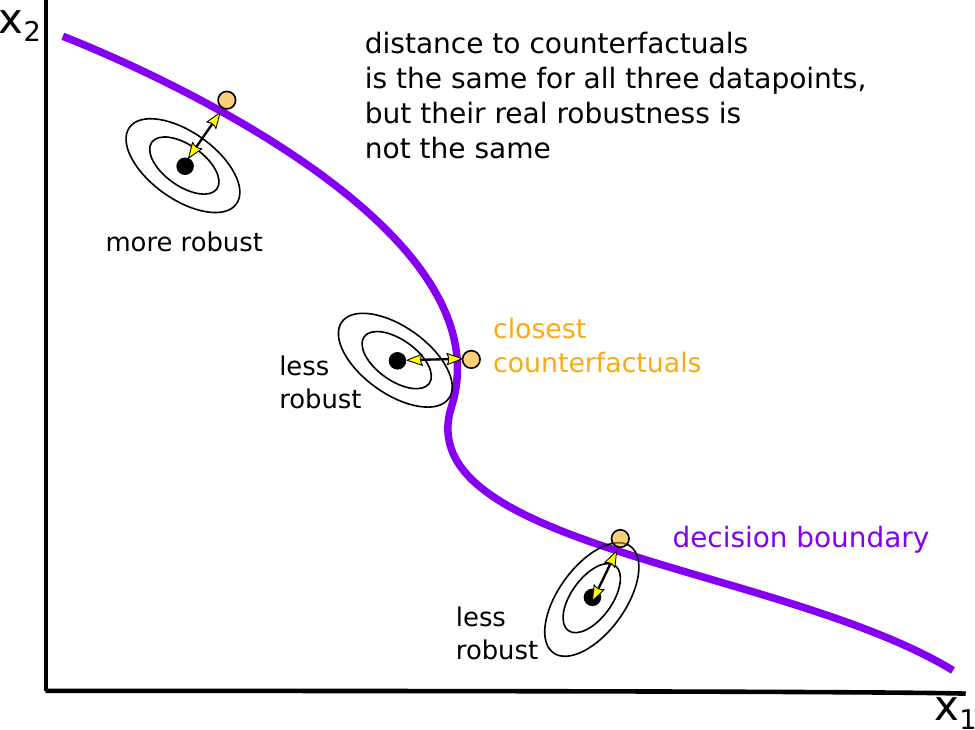}\caption{\label{fig:per-point-perturbations}Illustrations showing why the
distance to the closest counterfactual is an insufficient measure
for real-world-robustness.}
\end{figure}

The above discussed principle --- that there are many different situations
that can lead to the same probability of misclassification for a given
uncertainty distribution around the data point --- shows something important,
namely that the shortest distance to the decision boundary alone is
not enough for determining the robustness of a prediction. To revisit
fig.~\ref{fig:comparison-situations}: The right data point is much
closer to the decision boundary than the left one, and still both
have the same robustness in our sense. 
We thus have an intuitive explanation for why the distance to the closest counterfactuals
(or adversarial examples), as for example suggested in \cite{sharmaCERTIFAICommonFramework2020a},
is \emph{not} sufficient for determining robustness to perturbations
that could occur by chance in the real world. In fig.~\ref{fig:comparison-situations},
the closest counterfactual to the left data point would be much further
away than the closest counterfactual for the right data-point --- thus
suggesting a different robustness --- but in fact we have demonstrated above
that both points have the same robustness. This stems from the different
shapes of the decision boundary in the vicinity of the two points.
Another example is given in fig. \ref{fig:per-point-perturbations}a):
Here we have two data points (for simplicity again with both data points
having the same uncertainty), and the distance to their closest counterfactuals
(indicated as yellow dots) is the same for both points. Therefore,
the distance would indicate that they have the same robustness. However,
due to the different shapes of the decision boundary, the left point
actually is less robust than the right one. This is the case for all types of classifier that have curved decision boundaries (thus all non-linear classifiers). For completeness we point out that for linear classifiers (e.g. logistic regression), the distance to the closest counterfactual does not suffer from the above mentioned problems (but the issues discussed below are still relevant even for linear classifiers).

Another reason why the distance to the closest counterfactual is not
enough is depicted in fig.~\ref{fig:per-point-perturbations}b). Here
we have three data points, all with the same distance to the decision
boundary, and thus the same distance to the closest counterfactual
(orange points). In this example, also the shape of the decision boundary
is very similar in the vicinity of all three data points. However,
the uncertainty of the data points is not the same. All points have elliptical uncertainty distributions, but with different
orientation with respect to the decision boundary. Therefore, their
robustness against real-world perturbations is not the same --- the
leftmost point is more robust than the others. At a first thought,
one might be tempted to think that the latter problem might be solved
the following way: Given the density distribution $p\left(x_{t}\right)$,
we could compute the density at the location of the counterfactual
and use this as a robustness measure. In the example in fig. \ref{fig:per-point-perturbations}b)
this would indeed yield the desired result: The leftmost point would
be rated more robust then the two other points. However, if we return
to the example in fig. \ref{fig:per-point-perturbations}a), then
this approach would immediately fail: Here both points would be rated
equally robust --- as the counterfactuals lie on the same probability
lines --- but indeed we had shown above that in terms of real-world-robustness,
the left point is less robust than the right one. The opposite would
be the case in the example in fig. \ref{fig:comparison-situations}:
Here the density at the location of the closest counterfactual would
be higher for the right than for the left point --- but indeed the real-world
robustness is the same for both points.
This does of course not mean that counterfactuals/adversarial examples are useless in that respect --- it only means that they do not cover the whole picture.

\subsection{Analytical Solutions}
For combinations of certain types of classifiers and assumed uncertainties it is possible to find analytical solutions to Eq.~(\ref{eq:p-integral}). In the Appendix this is shown for two examples: 1) A linear classifier and 2) a non-linear rule-based model, where for both the uncertainty of the test-points is assumed to be a multivariate Gaussian distribution.

\subsection{Assessing Real-World-Robustness with MC Sampling}

In this section we describe a brute-force approach based on Monte-Carlo sampling that in principle can approximate the real-world-robustness of any black box classifier, on a test-point by test-point basis:

\begin{enumerate}
    \item{Build a function that returns discrete samples from $\hat{p}$.}
    \item{Draw $N$ samples from that function.}
    \item{Test the black box function $f$ with the new data points, and compute the fraction that leads to a different classification than the original data point.}
\end{enumerate}

For $N\rightarrow \infty$ this method should converge to the exact solution of Eq.~(4).

The challenge now lies in accurately describing the uncertainty $\hat{p}$. This needs to be done on an application-by-application basis and will require expert knowledge. Here we only outline some basic approaches that could be taken, in order to help practitioners.

\subsubsection{Continuous Features}
For continuous features, if appropriate for the application, the easiest approach would be to define $\hat{p}$ as a multivariate normal distribution. For $n$ features, this can be defined via a covariance matrix of shape $n \times n$. This can account for both uncorrelated and correlated uncertainty of the individual features.

\subsubsection{Categorical Features}
For categorical features one can work with transition matrices that describe transition-probabilities. If we have single categorical feature with $m$ possible values this would result in a (symmetric) $m \times m$ matrix. Each row (and each column) describes the transition probability of one categorical value to all other categorical values.

\subsubsection{Regression Problems}
The discussion in this paper focuses on classification problems --- as does most of the literature on robustness of machine learning algorithms. It is, however, possible to adapt our approach to regression problems as well. In this case, Eq.~(\ref{eq:binary}) would need to be adjusted with a threshold $\gamma$. If the difference between the regression output of the perturbed testpoint and the original testpoint is smaller than the threshold, then the output is considered unchanged, otherwise it is considered as changed:
\begin{equation}
f'\left(x_{t},x\right)=\begin{cases}
0 & \left|f\left(x\right)-f\left(x_{t}\right)\right|<\gamma \\
1 & \left|f\left(x\right)-f\left(x_{t}\right)\right|>\gamma
\end{cases}
\end{equation}
The rest remains the same as for classification problems.

\subsection{Assessing Real-World-Robustness in High-Dimensional Datasets}
The above described brute-force approach is bound to fail in settings with high-dimensional feature spaces due to the curse of dimensionality. In order to explore the whole area of interest around a test-point, a very huge number of samples would need to be drawn from $\hat{p}$, which is unfeasible.
Finding solutions to (part) of this problem is actually the topic of all work on certification and formal verification of neural networks mentioned in the introduction. However, to the best of our knowledge, no complete solution is known at the moment. Instead, the literature focuses only on providing guarantees for certain aspects of input perturbations.
Work on adversarial robustness of course also deals with these issues and partly provides solutions. However, as outlined in this paper, adversarial robustness is not necessarily the same thing as robustness to real-world-perturbations.
One interesting line of research would be to rephrase the question behind formal guarantees of neural networks: Instead of trying to find bounds within which perturbations do not change the prediction, one could attempt to formally guarantee that the probability of misclassification under a given (high-dimensional) perturbation strategy is below a certain probability threshold --- thus finding an upper bound for $P$ in Eq.~(\ref{eq:p-integral})

For certain applications, one could also use dimensionality-reduction to reduce the problem.
The number of input features in the input space does not always have to be the actual dimensionality of the input data, as features can be (highly) correlated. It could be that in some use-cases real-world-perturbations occur only --- or at least mainly --- along certain directions in feature space, which could coincide with the directions along which most variance occurs in the training or test data. Methods for dimensionality reduction --- such as Principal Component Analysis (PCA) --- could thus help in exploring the relevant space around test points. This would be possible through modelling real-world-perturbations not on the full input space, but on the leading principal components, or on feature found with stochastic games \citep{wickerFeatureGuidedBlackBoxSafety2018a}. This will, however, never work generically. For example, in some applications white noise will be a potential real world perturbation, but this will not be represented at all in the leading principal components.

Other potential approaches:

\begin{itemize}
    \item Try to use multiple different adversarial examples (e.g. not only the example closest to the original point, but the $n$ closest ones) as guidance towards the "interesting" direction(s) in feature space.
    \item For classifiers where the gradient of the output is available (e.g. for neural networks), it would be possible to compare the direction of the largest absolute gradient at the testpoint with the direction of highest probability in the assumed perturbation distribution around the test point. If the two directions are similar, then the robustness is lower.  This could for example be based on the approach used in the CLEVER score of \cite{wengEvaluatingRobustnessNeural2018a}. Such an approach would be, however, closer to an extension of adversarial robustness with information on perturbation distributions than a solution for finding real-world-robustness in our definition.
\end{itemize}

\section{Examples}
In the fist part of this section we compare adversarial robustness with real-world-robustness on classifiers trained on two commonly used datasets --- the Iris flower dataset \citep{andersonSpeciesProblemIris1936a} and the Ionosphere dataset \citep{sigillito1989classification}, both available in the UCI repository \citep{Dua:2019}.
In the second part, we turn to an example that is closer to a real world application, namely fog prediction at an airport, with known uncertainty of the input features.

\subsection{Simple Datasets}
The experiments we conduct are mainly for illustration purposes. On both datasets, a classifier is trained, and then robustness is estimated with different assumptions on how real-world perturbations could look like. We simply use two approaches as illustrations of the concept, they do not necessarily correspond to what would be real-world perturbations in applications that use classifiers based on these datasets.

For both datasets, we describe the expected uncertainty around the test-point as multivariate normal distributions, which are the same for each point (we thus assume that the uncertainty of input features is independent of their value). The multivariate normal distribution is described by a covariance matrix normalized to a trace of 1, and an overall perturbation scale, which in our example is a scalar.  We use two types of covariance matrices, an identity matrix (no covariance terms, all diagonal terms 1) and random covariance matrices (normalized to a trace of 1). The random covariance matrices are generated the following way:
\begin{equation}
    C_r=A^TA,
    \label{eq:rand-cov}
\end{equation}
where $A$ is a matrix whose entries are all independently drawn from a normal distribution with zero mean and unit variance. Eq.~(\ref{eq:rand-cov}) ensures that $C_r$ is a proper covariance matrix (symmetric and positive semi-definite).

Real-world-robustness in our definition is a probability, and therefore contained in $\left[0,1\right]$. The typical way of using adversarial examples as robustness measure is the distance $d_i$ between the test point $x_i$ and the closest adversarial example. This is a metric in $\left(0,\infty\right)$ (in contrast to real-world-robustness, the distance cannot be zero, therefore the interval has an open lower bound). In order to allow a better comparison, we rescale the distance to the range $\left(0,1\right]$, thus we define adversarial robustness as
\begin{equation*}
    r_i=\frac{d_i}{\max\left(d_{1,2,.....N}\right)}.
\end{equation*}

For each test point, $N=10000$ samples from the perturbation distribution around the test point are drawn for computing real-world-robustness.

\subsubsection{Iris Dataset}
As a very simple example we reduced the dataset to two classes (to make it binary). This results in an easy classification task.  The data is normalized to unit variance, then a random forest classifier with 10 trees and a max-depth of 3 is fitted on the training data (2/3 of dataset). Robustness computed on remaining 1/3 test points.
To compute the real-world-robustness we use the covariance matrices described above, whereas different perturbation scales are tested. counterfactuals/adversarial-examples are computed with the CEML library \citep{ceml}.

\begin{figure}
a)\includegraphics[width=0.3\textwidth]{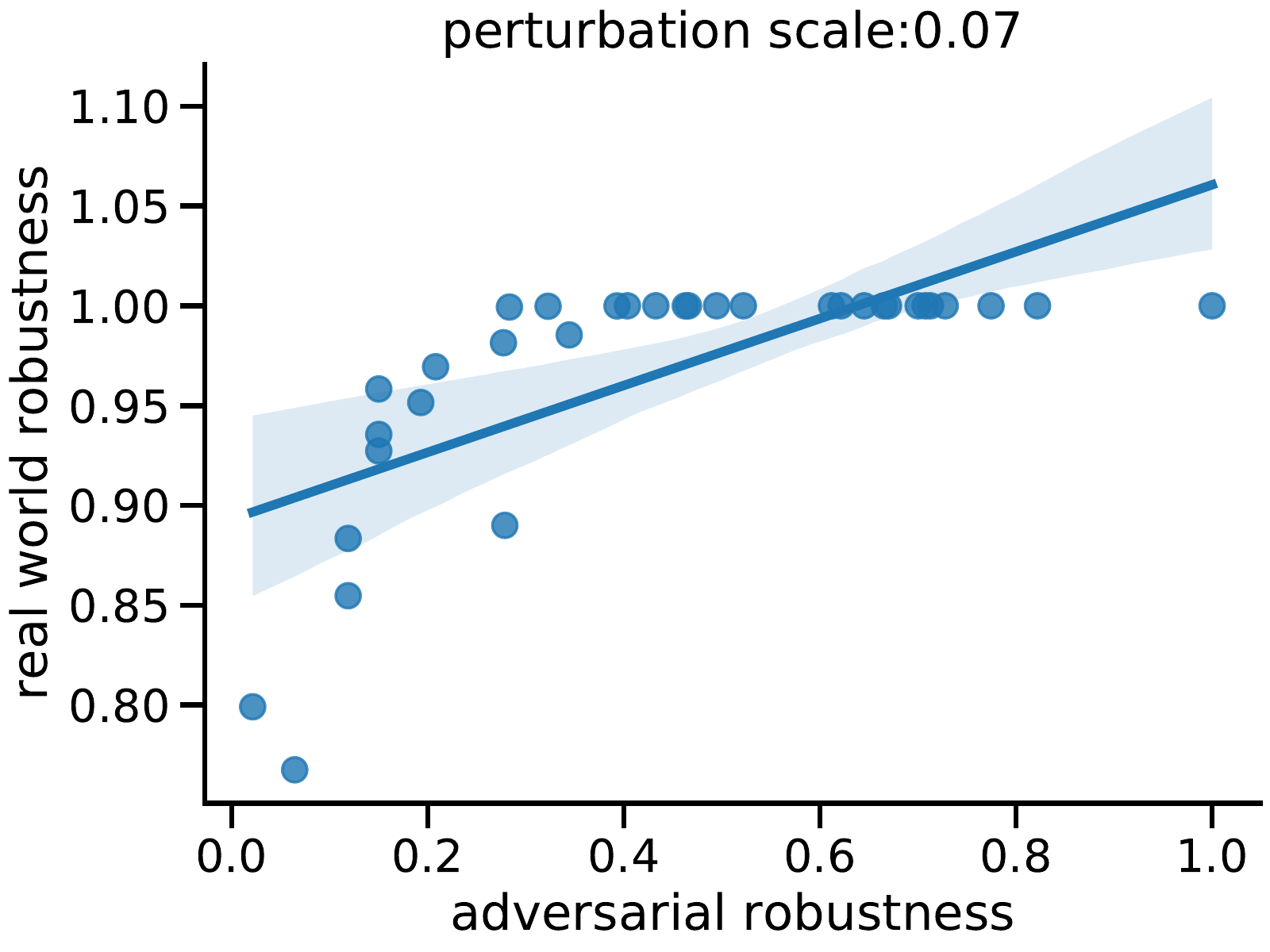}
b)\includegraphics[width=0.3\textwidth]{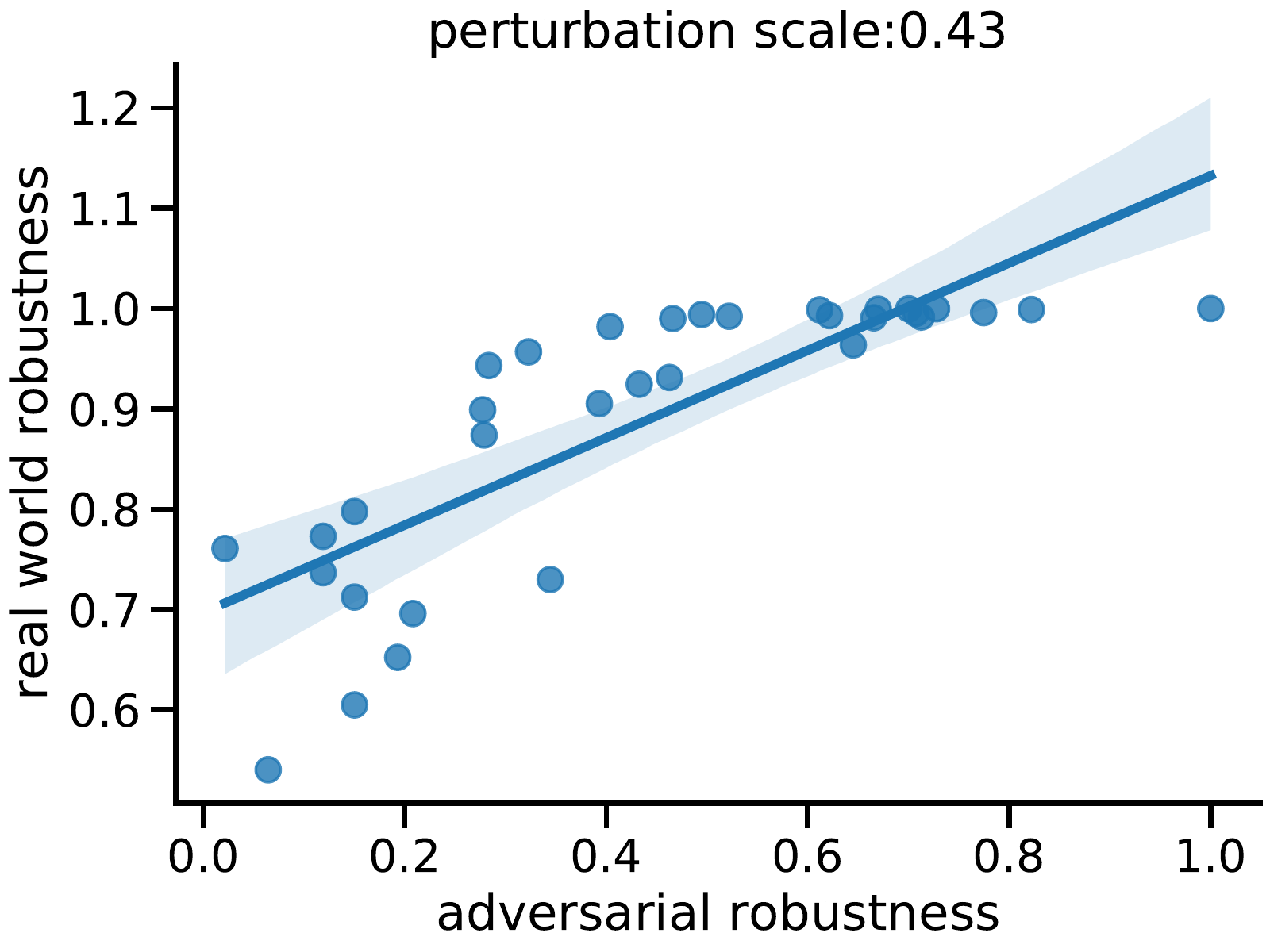} c)\includegraphics[width=0.3\textwidth]{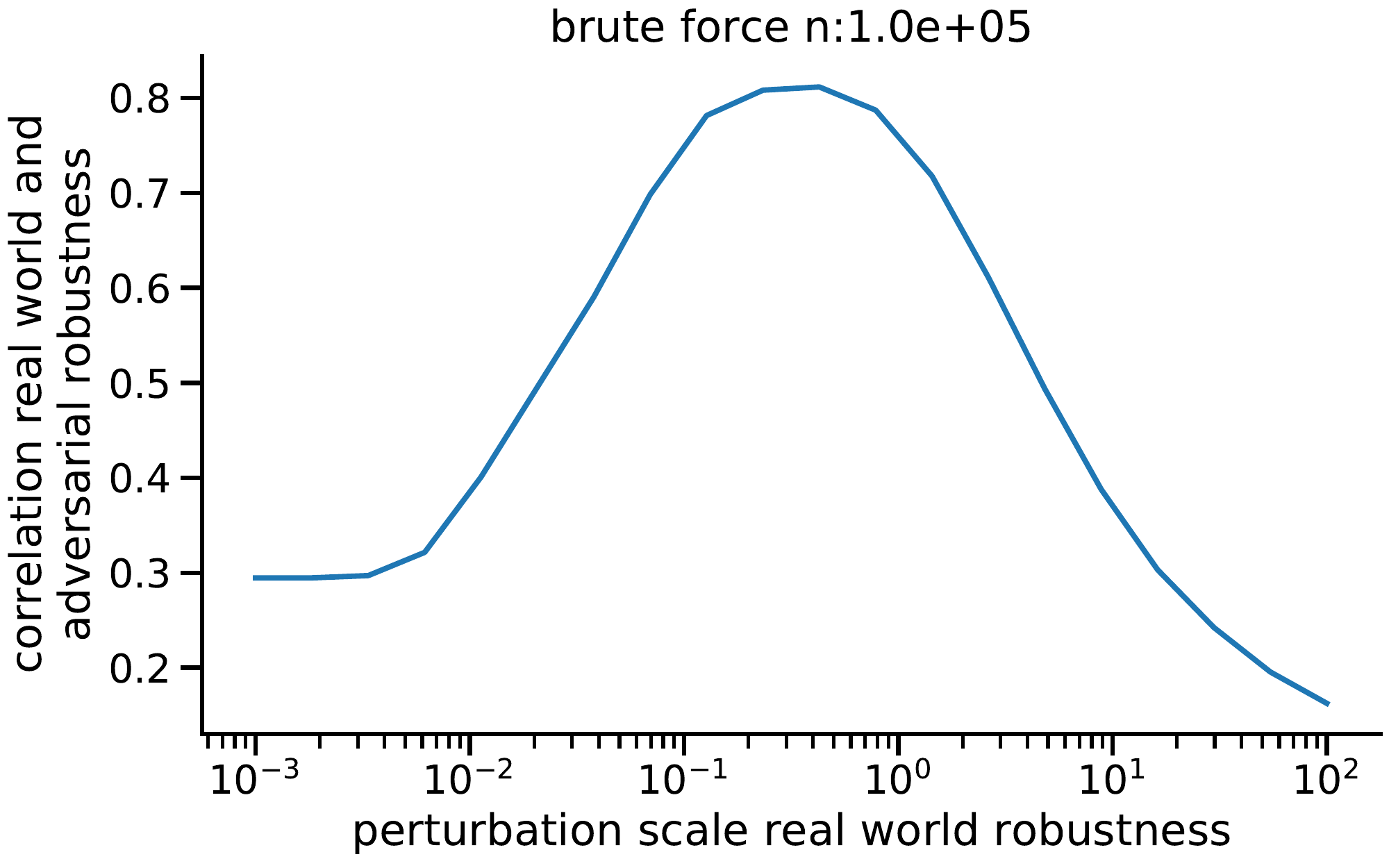}
\caption{\label{fig:iris}Example of adversarial robustness based on the Iris dataset (reduced to binary classification). a, b: Comparison between adversarial robustness and real-world-robustness with two different assumed scales of real world perturbations. c: Correlation between adversarial robustness and real-world-robustness for different real world perturbation scales.}
\end{figure}

\subsubsection{Ionosphere Dataset}
The ionosphere dataset has 34 features (2 categorical and 32 continuous), of which we removed the 2 categorical features. The task is binary classification. As model we use a neural network with 2 hidden layers with 256 neurons each and relu-activation functions.
As the CEML library had --- in contrast to on the Iris dataset --- convergence problems on this higher dimensional dataset, for the Ionosphere dataset we use the method by \cite{brendel2019accurate} as implemented in the Adversarial Robustness Toolbox \citep{nicolaeAdversarialRobustnessToolbox2019} to compute counterfactuals/adversarial-examples.

\begin{figure}
a)\includegraphics[width=0.3\textwidth]{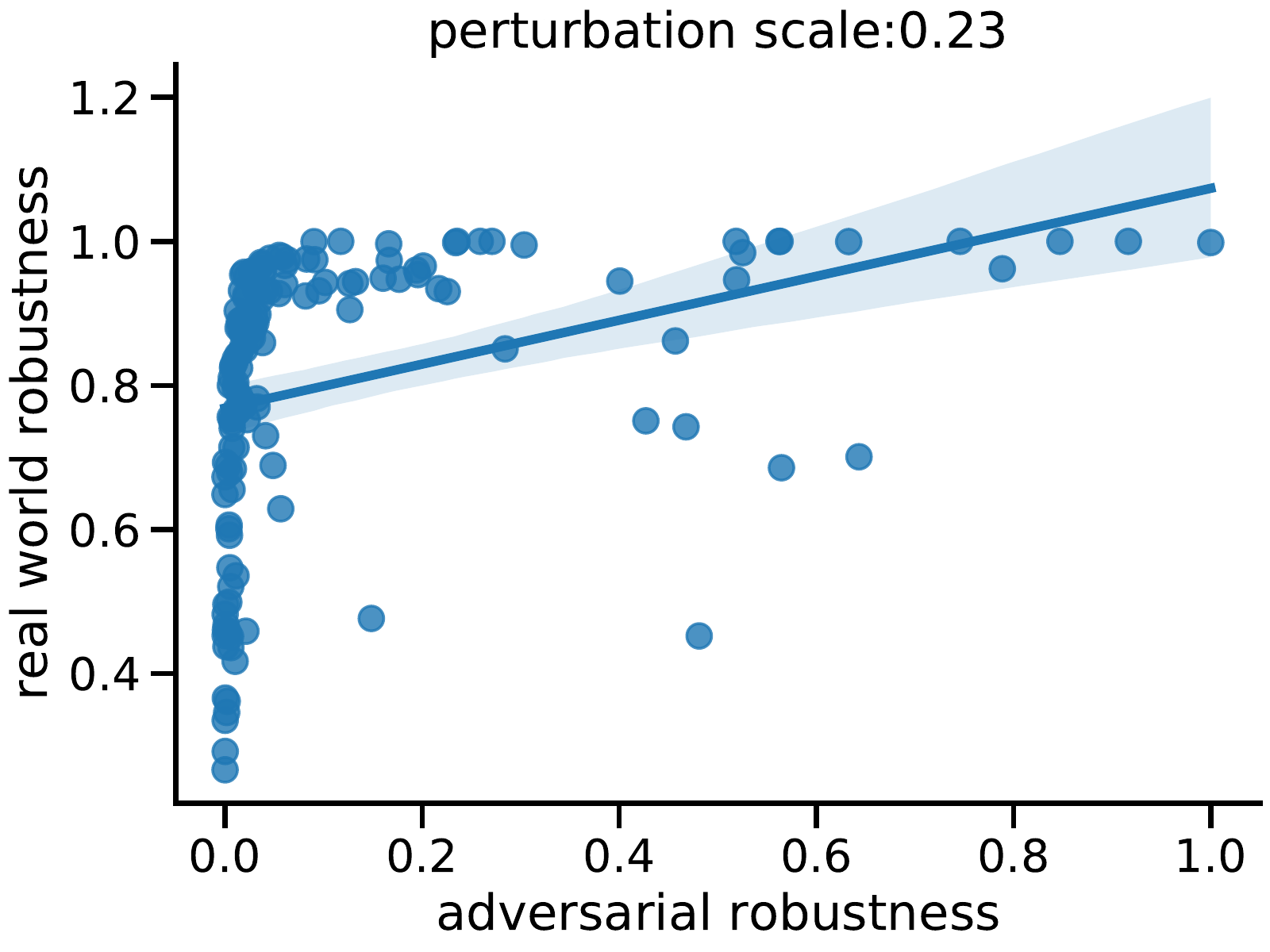}b)\includegraphics[width=0.3\textwidth]{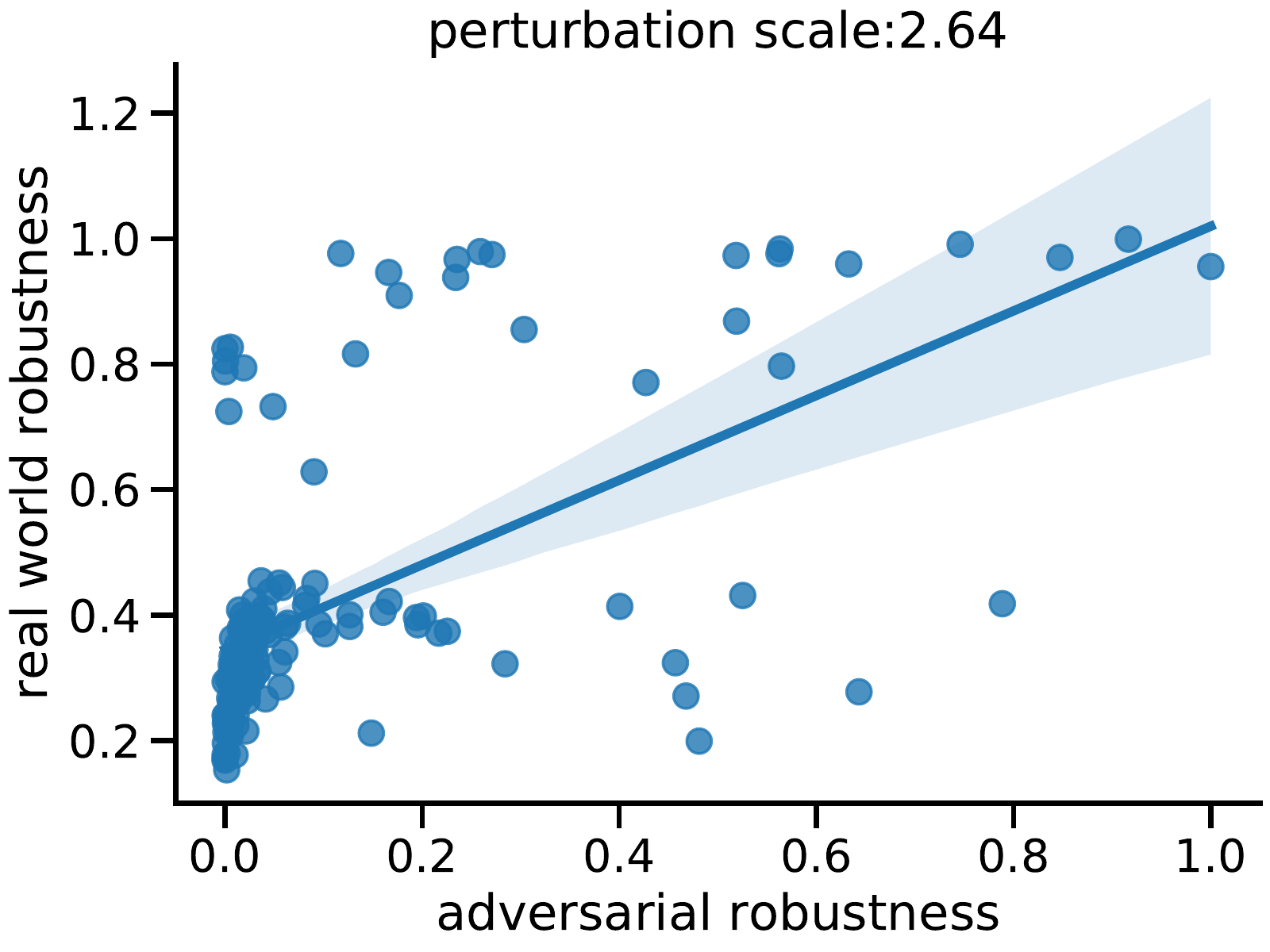}c)\includegraphics[width=0.3\textwidth]{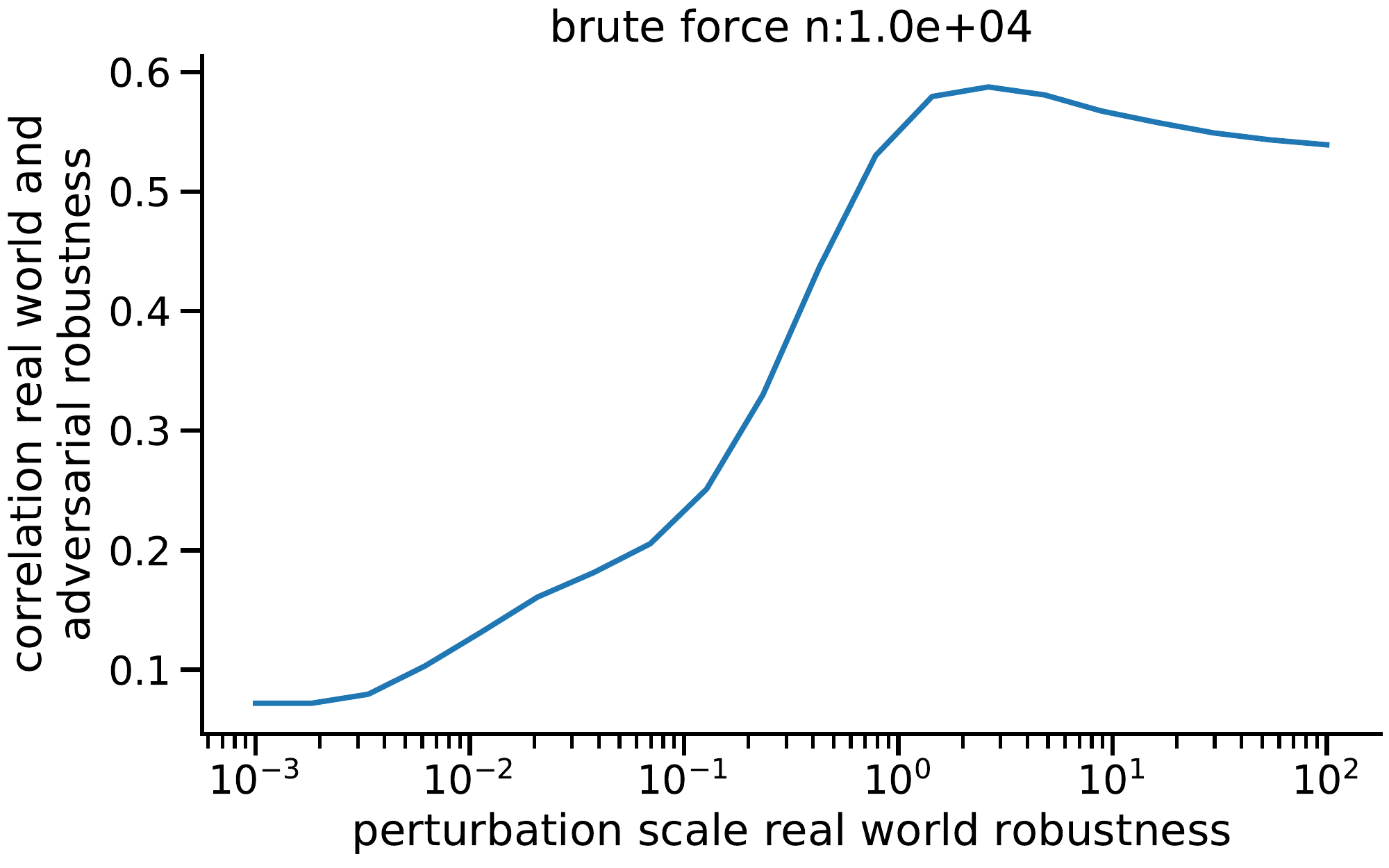}
\caption{\label{fig:ionos}Example on Ionoshpere dataset, comparing adversarial robustness and real-world-robustness. a,b: comparison between adversarial robustness and real-world-robustness with two different assumed scales of real world perturbations. c: correlation between adversarial robustness and real-world-robustness for different real world perturbation scales.}
\end{figure}

\subsubsection{Results}

Figures \ref{fig:iris} and \ref{fig:ionos} show the results with identity covariance matrix for Iris and Ionosphere dataset, respectively. The scatter plots (a,b) show two particular real world perturbation scales, panel c shows correlation between adversarial robustness and real-world-robustness for different assumed real world perturbation scales.  The two robustness measures clearly yield different results, showing that they are not measuring the same thing. The crucial thing is not that the correlation is not 1, but that the \emph{order} of the test points --- and therefore their ranking in terms or robustness --- is partly different. Another notable aspect is that the distribution of the two measures is different. Real-world-robustness, in contrast to the distance to the closest counterfactual, is dense close to a robustness of 1 (meaning probability of change in classification close to zero). This reflects the fact that if the test-point is already far away from the decision boundary, then it does not make a large difference for real-world-robustness it it would be even farther away. This is not correctly reflected by the distance to the closest counterfactual.

The perturbation scale where the correlation is highest indicates the scale of real perturbations (with the given cross-correlation across the features) at which the adversarial robustness measure works best in estimating real-world-robustness --- but still it does not represent real-world-robustness completely. Furthermore, it is unlikely that in a real application the expected perturbations correspond to exactly that scale, since the scale occurring in reality would be determined by the application, in contrast to the optimal scale, which is determined by the classifier.

To test whether the Monte Carlo approach for the computation of real-world-robustness has converged, for both datasets, the procedure has been repeated 20 times, and the correlation between the real-world-robustness obtained between all 20 runs computed. The correlation was larger than 0.999 for all pairs of runs, showing that it has converged (not shown).

\begin{figure}
    \centering
    a)\includegraphics[width=0.45\textwidth]{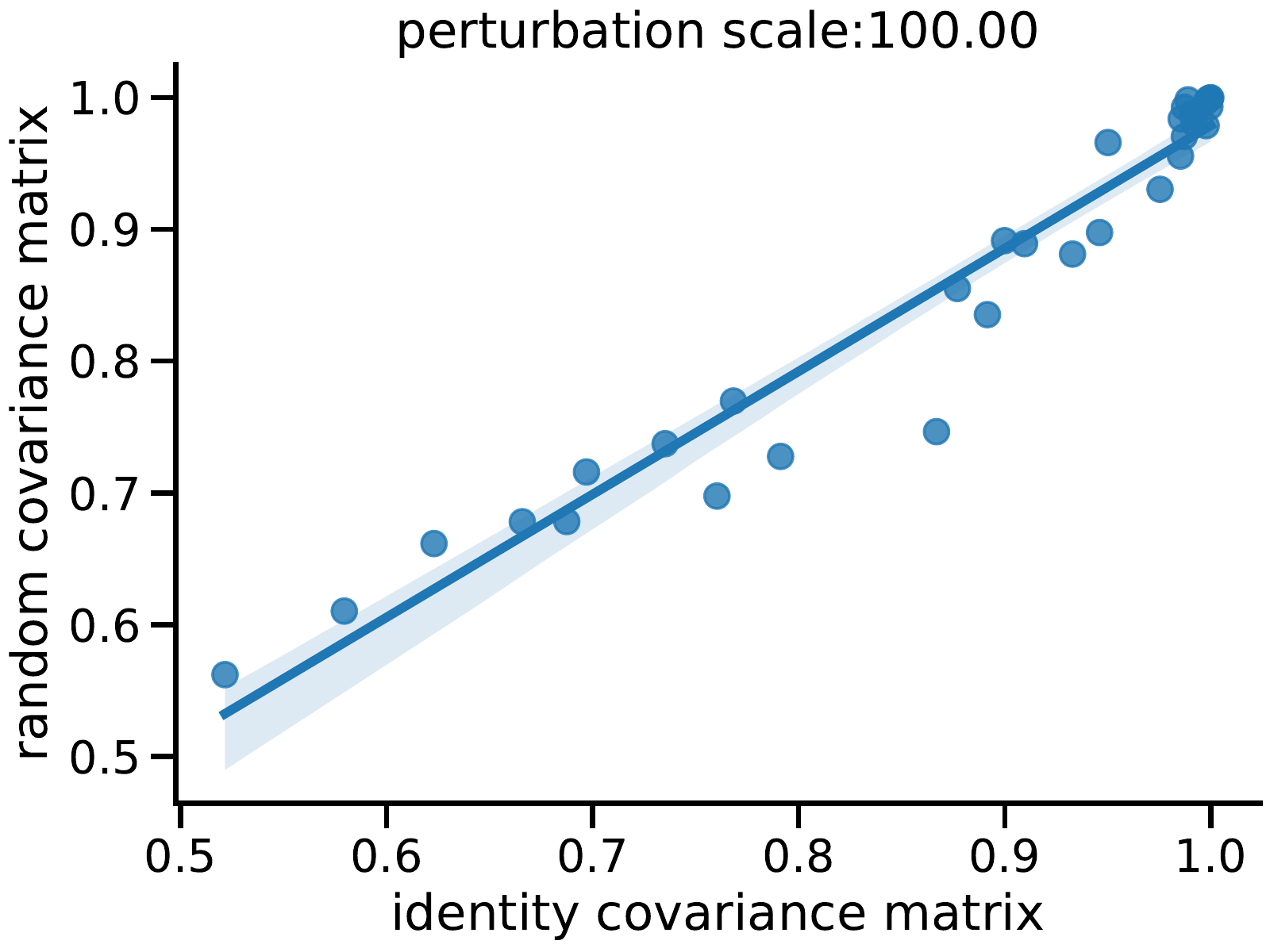}b)\includegraphics[width=0.45\textwidth]{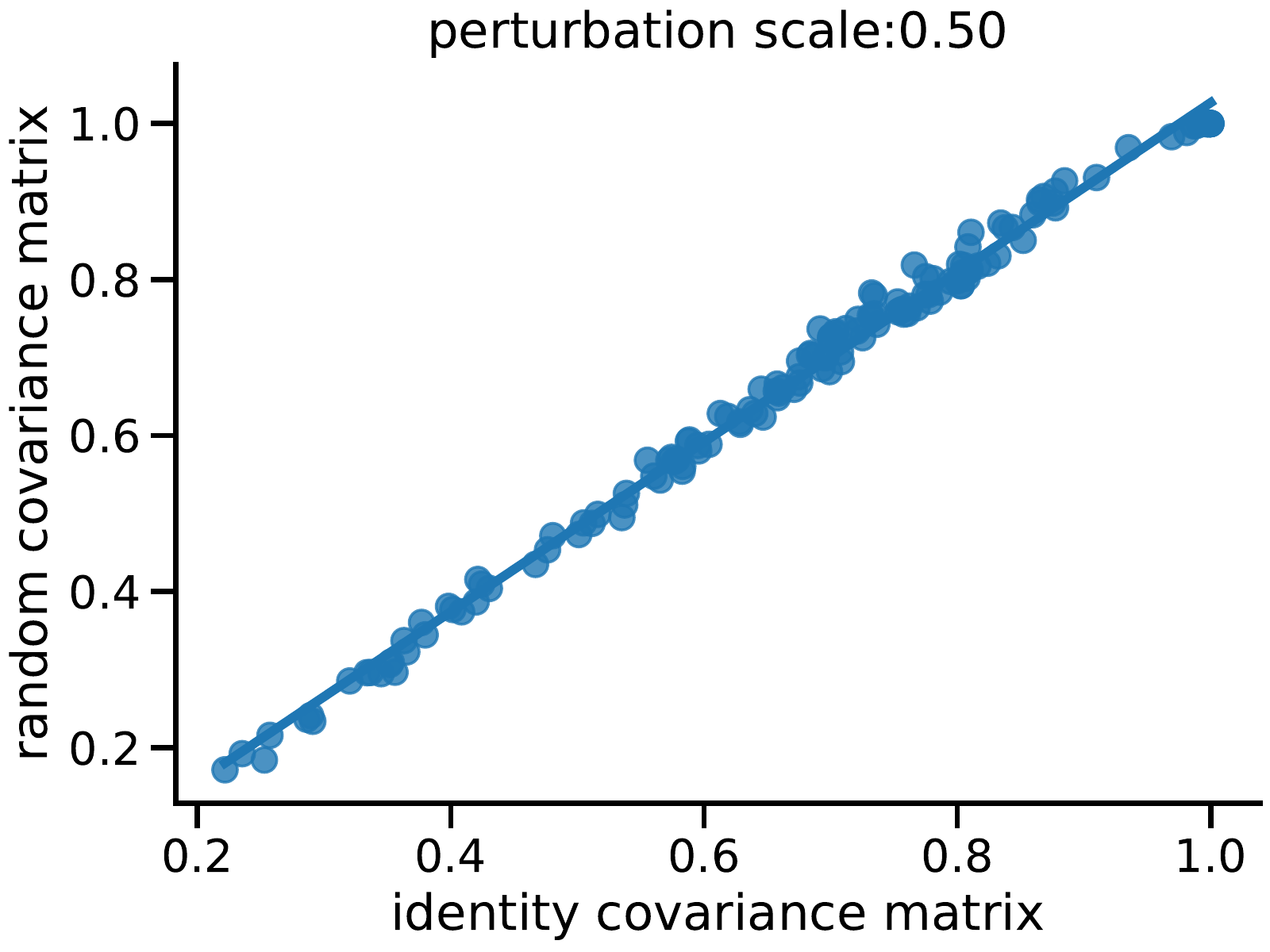}
    \caption{Real-world-robustness for uncertainty modelled with no covariance (identity matrix as covariance matrix) compared to real-world-robustness for uncertainty modelled with random covariance matrix (with same trace) for iris (a) and ionosphere (b) dataset.}
    \label{fig:cov-novoc}
\end{figure}

Finally a comparison between using identity matrices and random covariance matrices for real-world-robustness is show in fig. \ref{fig:cov-novoc}. For the ionosphere dataset, the results are very similar, but for the Iris dataset, assuming different perturbation distributions around the test-points leads to different robustness estimations. This shows that knowledge of the exact expected perturbations in the real-world is important for accurately estimating robustness not only from a theoretical example, but also for a real trained classifier.

\subsection{Fog Prediction}
The tests on the simple datasets provided insights into how assumptions about test point uncertainty influence real-world-robustness.
We now test the robustness calculation with a model designed and trained for a real-world application, for which reliable information of the test-point uncertainty is available, namely fog prediction at an airport based on coarse weather data. Predictors are 10 weather variables from the geographically closest gridpoint of the ERA5 reanalysis dataset \citep{hersbach2020era5}, which represents coarse weather information assimilated on a 25x25km grid. The target for the prediction is whether visibility as observed locally at the airport of Graz, Austria, is below or above 10km (value 60 in the Synop-coded data). Visibility values are taken from the local weather station of the Austrian Weather Service (GeoSphere Austria). 
We use hourly data spanning the year 2022. Due to the high temporal autocorrelation in the data, we use a block-approach for the train-test-validation-split. We use the first 2 weeks, as training, week 3 as validation, and week 4 as testing data, and then continue this way (week 5-6 raining, week 7 validation, week 8 test, and so on).
Based on \cite{ortega2019}, who tested different models for a related visibility prediction task, we chose as model a 2-layer Neural Network. For each point and each timestep, ERA5 contains both a best guess, and an uncertainty estimation based on a 10-member ensemble. For the training, we use the best-guess value (as it is done in most applications). For computing the \emph{real-world-robustness}, we can take advantage of the ensemble. For each timestep, the reanalysis ensemble consists of 10 different members, whereas each member has connected values for all variables (thus each member consists of one vector for all variables). This allows to not only estimate the uncertainty of each variable at each datapoint, but we can additionally estimate the covariance terms between the uncertainties of the individual variables. Thus, for each datapoint in the test set, we can compute the covariance matrix that describes the uncertainty of that individual test point.
We compare the obtained real-world-robustness with adversarial robustness computed with the DeepFool \citep{DBLP:journals/corr/Moosavi-Dezfooli15} algorithm as implemented in the adversarial robustness toolbox.

\begin{figure}
   a)\includegraphics[width=0.45\textwidth]{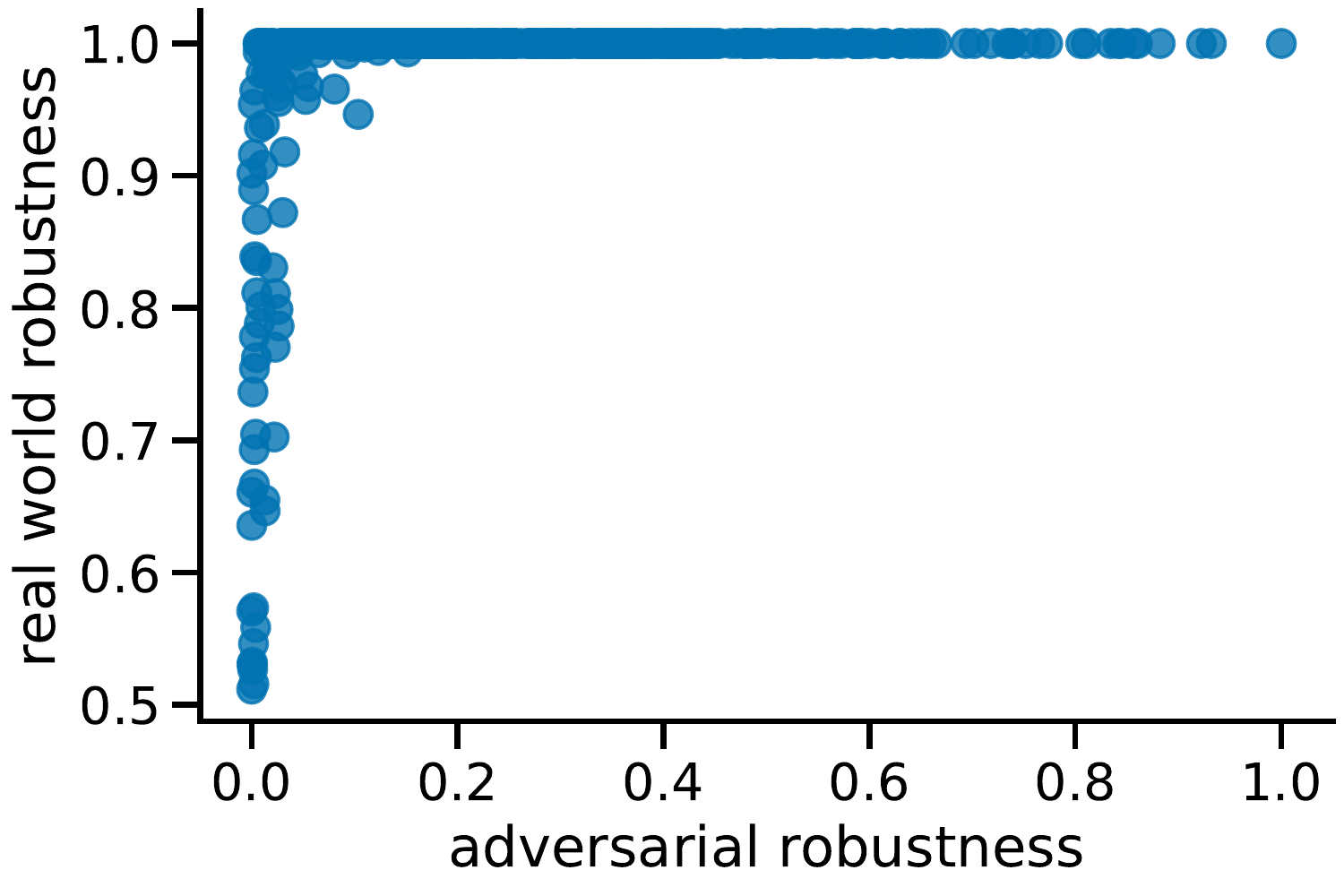}
     b)\includegraphics[width=0.45\textwidth]{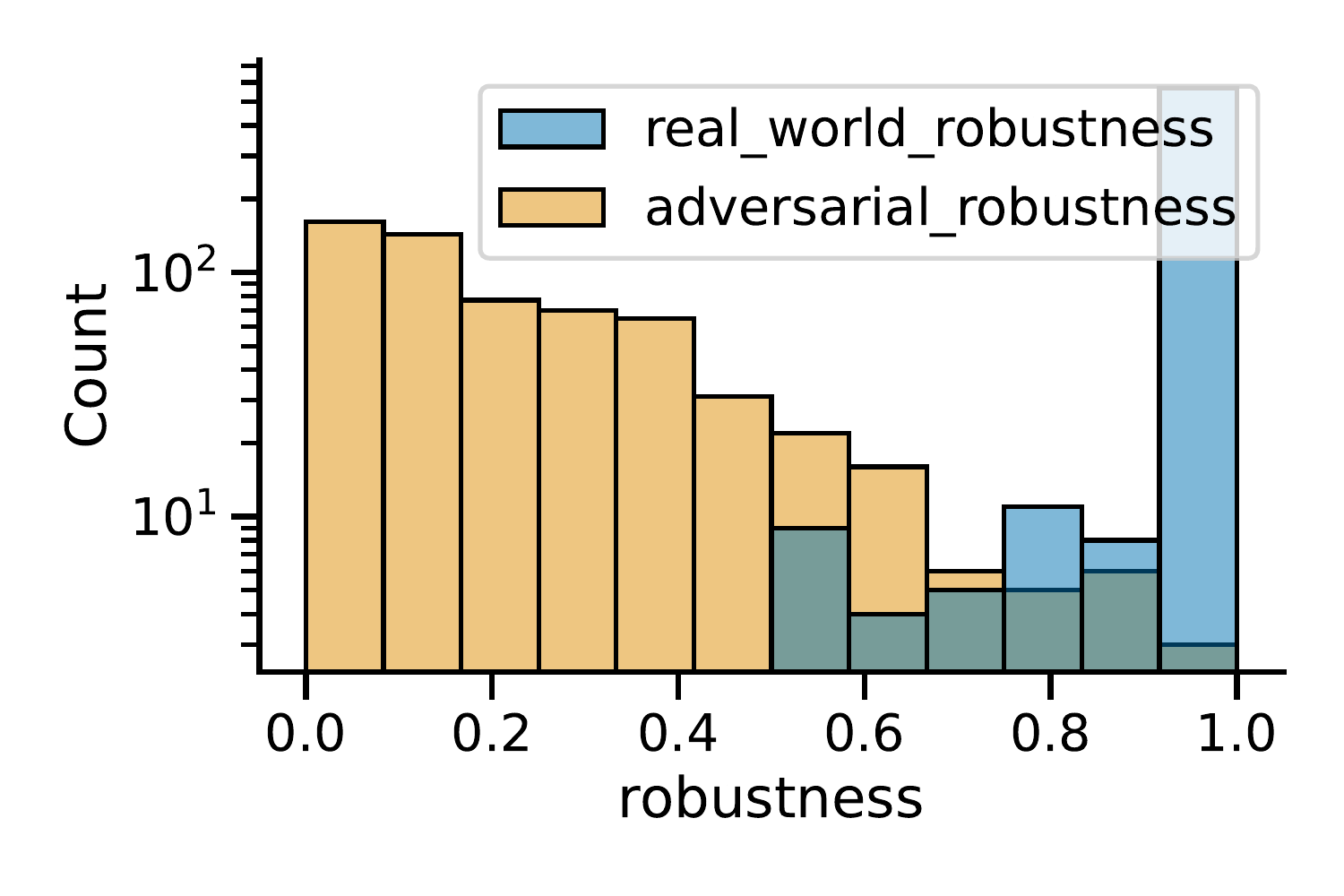}
    \caption{Adversarial robustness vs real-world-robustness of test points of a fog-classifier for Airport Graz (a) and histogram of both (b). Higher values denote higher robustness. Note the logarithmic scale on the y-axis in (b).}
    \label{fig:fog}
\end{figure}

As for the simple datasets, the convergence of the computation of real-world-robustness has been ensured by repeating the computation 20 times, and the minimum correlation was larger than 0.999 (not shown).
The results are depicted in fig. \ref{fig:fog}. There is a large number of test points for which \emph{real-world-robustness} is nearly 1. Thus for these points, the prediction is basically independent of the exact value of the input parameters within the conditional uncertainty at that point. On the other hand, there also cases where the robustness is lower, but interestingly, for no point it is lower than 0.5. Thus for no prediction is it more likely than not that the predictions change under uncertain input values. Note that this is not per-se true for every good binary classifier, as in the training, the uncertainty information of the individual points was not available. Also note this probability is not the probability of the classification, given the training data, but it is the probability of the classification, given the classifier and the test point uncertainty. 
This result clearly shows the value of \emph{real-world-robustness}, as from adversarial robustness or similar measures it would not be possible to deduce this information.
\section{Discussion}
One of the shortcomings when using counterfactuals for approximating real-world-robustness in our definition is that real-world-perturbations are not necessary equally likely along all directions of the input space. This could at least partly be adopted into counterfactuals via an adaption of the distance function in eq. \ref{eq:CF}. This would, however, not alleviate the problem that the classification boundary might have a different distance to the test point in different directions, and thus suffers from the same shortcomings as the approach of using the density of the input perturbations at the location of the counterfactuals mentioned in sec. \ref{sec:Why-counterfactuals-are-not-enough}.

Methods for finding counterfactuals only approximately solve the minimization problem in \ref{eq:CF}. In our discussion of the short-comings, we assumed \emph{perfect} counterfactuals. However, counterfactuals actually found by counterfactual methods will deviate from perfect counterfactuals, and thus are even less suited for approximating real-world-robustness.

\section{Conclusion}

Determining the robustness of individual predictions of ML classifiers is an important problem in  many applications, for example in settings where predictions of the model could potentially be challenged (e.g. in insurance settings), or where the uncertainty of the inputs change from prediction to prediction (e.g. in weather forecast settings).
In this paper we discussed how to assess robustness of individual predictions of machine learning classifiers to real-world-perturbations.  We defined \emph{real-world-perturbations} as perturbations that can occur ``naturally'' in a given real world application, for example through noise or measurement errors. These perturbations can be different for each input sample. This approach is different from adversarial attack scenarios, in which the threat scenario is that an adversary modifies input data of the classifier in order to  change its classification. Additionally, we differentiated it from robustness against distribution shifts, formal certification of neural networks and the problem of testing sets that correspond to testing ``in the wild''. 
Finally, we discussed the problem of estimating the robustness for individual predictions compared to the robustness of a classifier in general.
Leveraging the ideas behind perturbation robustness and common corruptions from \cite{hendrycks2019benchmarking}, we defined real-world-robustness in a mathematical way, and 
showed theoretically why measures for adversarial robustness are not necessarily a valid measure for real-world-robustness, as they measure something different than our definition of real-world-robustness. 
We further proposed that real-world-robustness can --- at least for low to medium dimensional problems --- be estimated with Monte Carlo sampling, which requires careful modelling of perturbations that are expected for a certain application. This can only be done with expert knowledge.

We empirically tested our approach on  classifiers trained on two widely used ML-datasets (the Iris flower dataset and the Ionosphere dataset) as well as an actual application predicting fog at an airport, and compared the results to robustness estimated with adversarial examples. The results were  different, showing that the two approaches do indeed measure two distinct things. While it has been shown before that in terms of general robustness of a classifier, high robustness to adversarial examples does not guarantee robustness to naturally occurring perturbations \citep{hendrycks2019benchmarking}, we showed that also the ranking of robustness between different samples is not the same for adversarial robustness as for robustness to real-world-perturbations.

Our result shows that claims that robustness scores based on adversarial examples / counterfactuals are generic robustness scores (such as in  \cite{sharmaCERTIFAICommonFramework2020a}) are not completely true.

An inherent limitation in our proposed way for estimating real-world-robustness is that it works only of the dimensionality of the input space is not too high. We discussed several ideas on how this could be dealt with. For certain problems, it should be possible to use dimensionality reduction techniques, but for general problems we cannot offer any solution yet. 
We suspect that the research fields of adversarial robustness and formal guarantees for neural networks can provide valuable input to this point. Future research should therefore, in addition to adversarial attack scenarios, also focus on how to determine \emph{real-world-robustness} for high-dimensional problems.

Finally, in this paper, we mainly dealt with black-box classifiers (except in the section
on analytical solutions).
In cases where the classifier is known, and more specifically the decision boundary(s) is known (such as in decision trees), it would also be possible to use numerical integration to compute \emph{real-world-robustness} (eq. \ref{eq:p-integral}). That would be done via numerically integrating $p$ over the region(s) where the classification is the same as for the test point, and should in principle also work for high-dimensional datasets. This possibility was not explored in this paper.


\section{Code and Data Availability}
The datasets we used are all publicly available. The Iris and the Ionosphere dataset can be obtained from the UCI repository (\url{https://archive.ics.uci.edu/ml/index.php}). The ERA5 Data can be obtained from Copernicus (\url{https://cds.climate.copernicus.eu/}). The station data from GeoSphere Austria can be obtained from the GeoSphere Austria Data Hub (\url{https://data.hub.zamg.ac.at/dataset/synop-v1-1h}). All code developed for this study and links to the datasets are available in the accompanying repository (repository will be published with the camera-ready version of this paper).

\acks{We thank Jo\~{a}o de Freitas for interesting discussions.

This work was partly supported by the ``DDAI'' COMET Module within
the COMET -- Competence Centers for Excellent Technologies Programme, funded by the Austrian Federal Ministry (BMK and BMDW), the Austrian Research Promotion Agency (FFG), the province of Styria (SFG) and partners from industry and academia. The COMET Programme is managed by FFG.}


\newpage

\appendix
\renewcommand\thefigure{\thesection.\arabic{figure}}   
\setcounter{figure}{0} 
\section{Analytical solutions}
\label{app:analytical-solutions}



The analytical solution of real-world-robustness is presented here
for two simple cases: A linear decision boundary, and a simple rule
based model. Both are binary classifiers with 2-dimensional feature
space.

\subsection{Linear decision boundary}

A linear decision boundary in 2-dimensional feature space can be described
as
\begin{equation*}
y\left(x_{1},x_{2}\right)=w_{1}x_{1}+w_{2}x_{2}+b,
\end{equation*}
with the corresponding binary classifier function
\begin{equation*}
f\left(x_{1},x_{2}\right)=\begin{cases}
0 & y<1/2\\
1 & y>1/2
\end{cases}
\end{equation*}

For this classifier we can compute analytical solutions both for the
distance to the closest counterfactual as well as for our definition
of real-world-robustness.

The closest counterfactual for a test-point $(x_{1},x_{2})$ is obtained
by minimizing the distance from $(x_{1},x_{2})$ to the decision boundary
$1/2=w_{1}x_{1}+w_{2}x_{2}+b$. If we use euclidean distance, this
results in 

\begin{equation*}
x_{cf1}=-\frac{2w_{2}x_{2}+2b+x_{1}-1}{2w_{1}-1}
\end{equation*}
\begin{equation*}
x_{cf2}=-\frac{2w_{1}x_{1}+2b+x_{2}-1}{2w_{2}-1}
\end{equation*}
The distance $d_{cf}$ to the counterfactual is given by 
\begin{equation*}
d_{cf}=\sqrt{\left(x_{1}-x_{cf1}\right)+\left(x_{2}-x_{cf2}\right)}.
\end{equation*}
We assume multivariate Gaussian uncertainty $p\left(x_{1},x_{2},x_{1}',x_{2}'\right)$
with covariance matrix $\sum=\left(\begin{array}{cc}
\sigma_{1}^{2} & 0\\
0 & \sigma_{2}^{2}
\end{array}\right)$ given by

\begin{equation}
p\left(x_{1},x_{2},x_{1}',x_{2}'\right)=\frac{1}{2\sigma_{1}\sigma_{2}}\exp\left[-\frac{\left(x_{1}'-x_{1}\right)^{2}}{2\sigma_{1}^{2}}-\frac{\left(x_{2}'-x_{2}\right)^{2}}{2\sigma_{2}^{2}}\right]\label{eq:gauss-uncertainy}.
\end{equation}

The real-world-robustness $P_{r}$, given by Eq.~(\ref{eq:p-integral}) in the main paper, can then be computed
as $1-P$, where $P$ is obtained by integrating $p$ over the
region of $\mathbb{R}^{2}$ where $f\left(x+x'\right)\neq f\left(x\right)$.
Alternatively, $P_{r}$ can be directly computed via $p$ over the
region of $\mathbb{R}^{2}$ where $f\left(x+x'\right)=f\left(x\right)$:
\begin{equation}
P_{r}\left(x_{1},x_{2}\right)=\int\int_{f\left(x+x'\right)=f\left(x\right)}p\left(x_{1},x_{2},x_{1}',x_{2}'\right)dx_{2}'dx_{1}'\label{eq:rwrb-integral}.
\end{equation}

This equation can be solved via a rotation of the coordinate system such that
one of the rotated coordinate axes is parallel to the decision boundary.

For illustration purposes here we assume a classifier with $w_{1}=0$
and $b=0$, in which case the decision boundary is parallel to the
$x_{1}$-axis.

\begin{figure}

\includegraphics[width=0.5\textwidth]{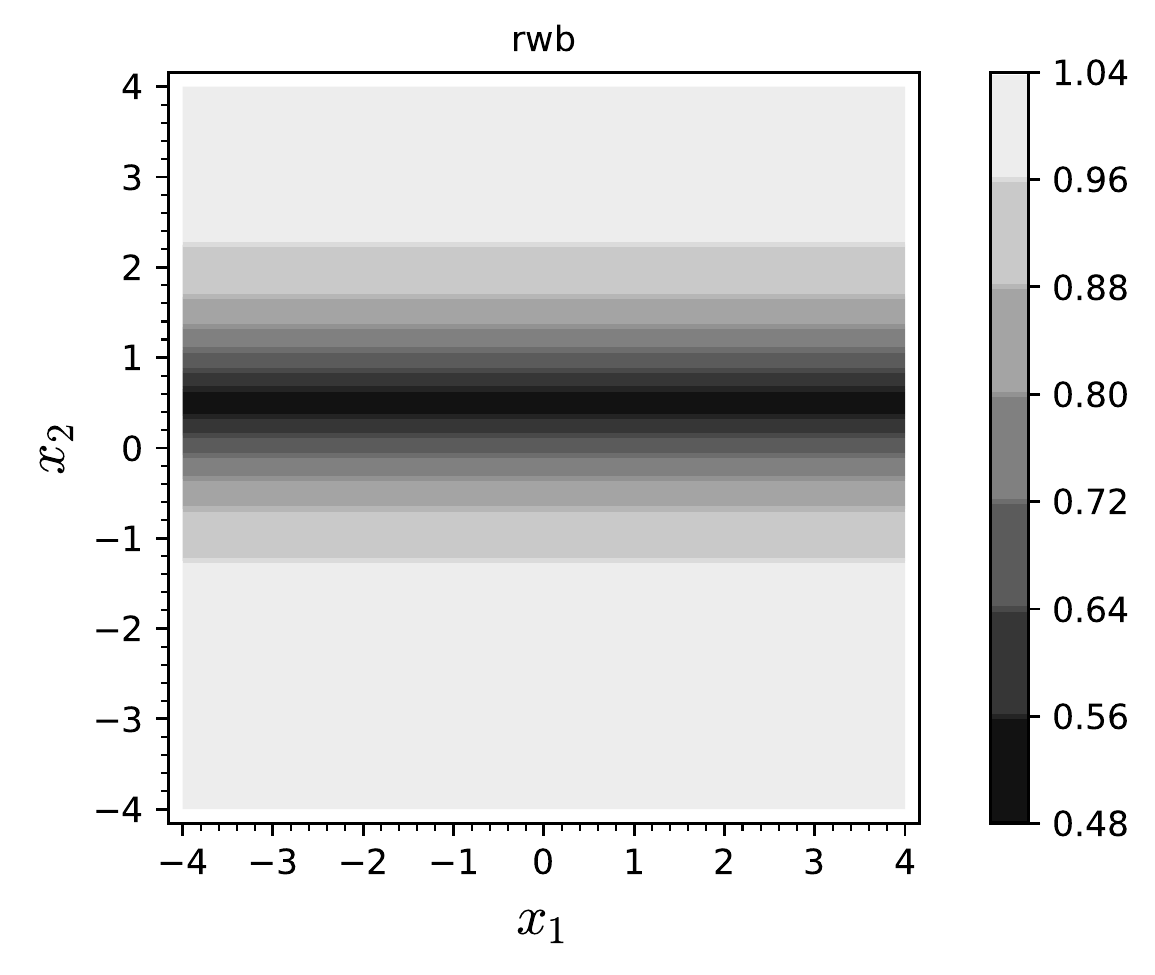}\includegraphics[width=0.5\textwidth]{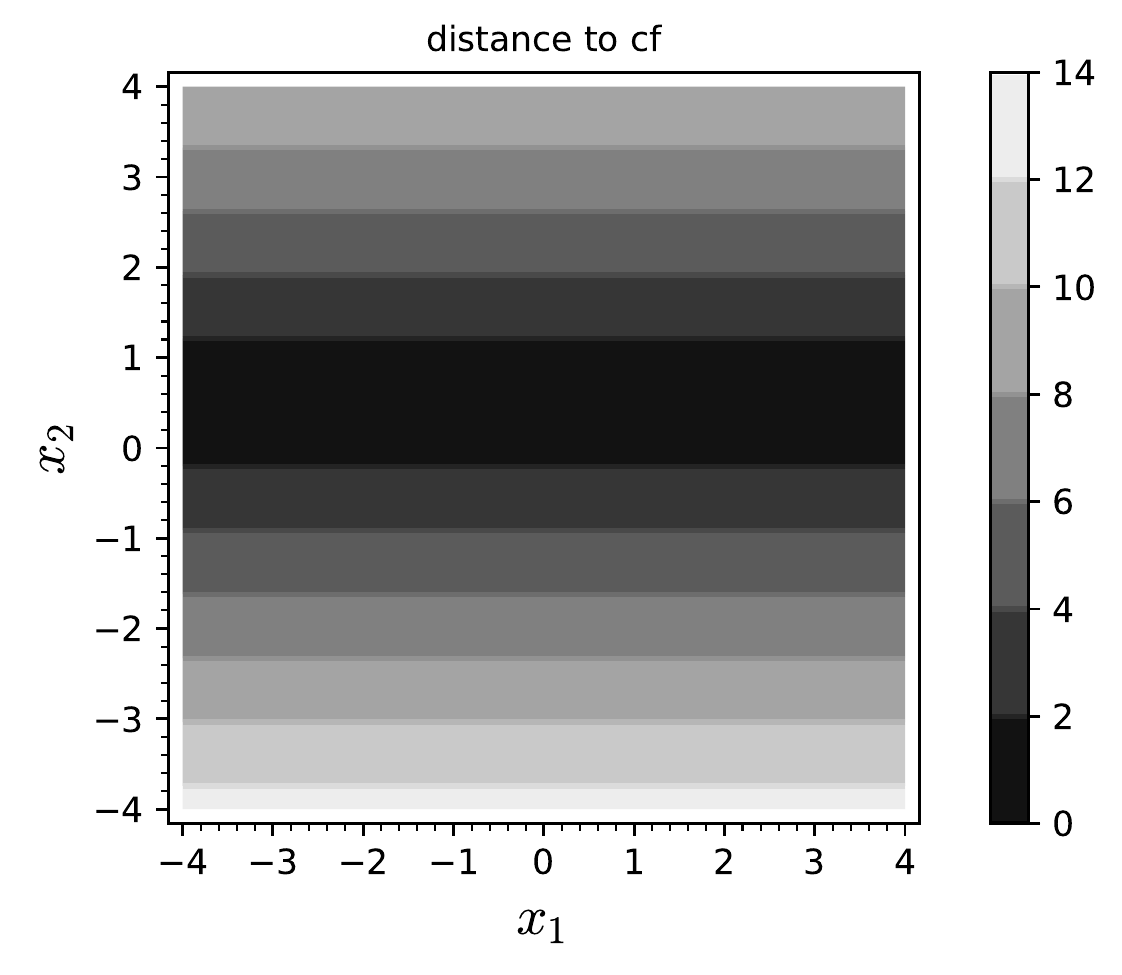}\caption{\label{fig:analytical-solution-linear}Analytical solution of real-world-robustness
(left) and distance to closest counterfactual (right) for a 2-dimensional
linear decision boundary with $w_{1}=0,w_{2}=1,b=0$ and Gaussian
test-point uncertainty with $\sigma=1$. Here the distance to the closest counterfactual is not normalized.}
\end{figure}

In that case Eq.~(\ref{eq:rwrb-integral}) simplifies to 
\begin{equation*}
P_{r}\left(x_{1},x_{2}\right)=\begin{cases}
\stackrel[-\infty]{\infty}{\int}\stackrel[\frac{1}{2w_{2}}]{\infty}{\int}p\left(x_{1},x_{2},x_{1}',x_{2}'\right)dx_{2}'dx_{1}' & x_{2}>\frac{1}{2w_{2}}\\
\stackrel[-\infty]{\infty}{\int}\stackrel[-\infty]{\frac{1}{2w_{2}}}{\int}p\left(x_{1},x_{2},x_{1}',x_{2}'\right)dx_{2}'dx_{1}' & x_{2}<\frac{1}{2w_{2}}
\end{cases}
\end{equation*}
with the analytical solution
\begin{equation}
P_{r}\left(x_{1},x_{2}\right)=\begin{cases}
\frac{1}{2}\,\operatorname{erf}\left(\frac{2\,\sqrt{2}w_{2}x_{2}-\sqrt{2}}{4\,\sigma_{2}w_{2}}\right)+\frac{1}{2} & x_{2}>\frac{1}{2w_{2}}\\
-\frac{1}{2}\,\operatorname{erf}\left(\frac{2\,\sqrt{2}w_{2}x_{2}-\sqrt{2}}{4\,\sigma_{2}w_{2}}\right)+\frac{1}{2} & x_{2}<\frac{1}{2w_{2}}
\end{cases}
\end{equation}
where $\operatorname{erf}$ is the error function.
Example plots for both robustness metrics are shown in fig.~\ref{fig:analytical-solution-linear}. Both are independent of $x_1$, and decline steadily with larger distance along $x_2$ from the decision boundary along. Distance to the closest counterfactual declines linear with distance (per definition of the used distance metric), real-world-robustness declines exponentially (because gaussian uncertaity is assumed).

\subsection{Non-linear decision boundary}

As an example of a non-linear classifier for which both robustness metrics
can be analytically computed we use the following decision function 

\begin{equation*}
f\left(x_{1},x_{2}\right)=\begin{cases}
1 & x_{1}>a_{1}\:\text{and}\:x_2>a_{2}\\
0 & \text{else}
\end{cases}
\end{equation*}
which forms a decision boundary in 2d that looks like a right angle,
where the tip of the edge is at $\left(a_{1},a_{2}\right).$ 
This could for example originate from a simple rule-based model, or
a decision tree with depth 2.

Distance to the closest counterfactual (i.e. distance to the decision
boundary) is given by
\begin{equation*}
d_{cf}\left(x_{1},x_{2}\right)=\begin{cases}
\min\left(\sqrt{\left(x_{1}-a_{1}\right)^{2}},\sqrt{\left(x_{2}-a_{2}\right)^{2}}\right) & x_{1}>a_{1}\:\text{and}\:x_2>a_{2}\\
\sqrt{\left(x_{2}-a_{2}\right)^{2}} & x_{1}>a_{1}\:\text{and}\:x_2<a_{2}\\
\sqrt{\left(x_{1}-a_{1}\right)^{2}} & x_{1}<a_{1}\:\text{and}\:x_2>a_{2}\\
\sqrt{\left(x_{1}-a_{1}\right)^{2}+\left(x_{2}-a_{2}\right)^{2}} & x_{1}<a_{1}\:\text{and}\:x_2<a_{2}
\end{cases}
\end{equation*}
The real-world-robustness thus follows as
\begin{equation*}
P_{r}\left(x_{1},x_{2}\right)=\begin{cases}
\begin{split}
\stackrel[a_{2}]{\infty}{\int}\stackrel[a_{1}]{\infty}{\int}p\left(x_{1},x_{2},x_{1}',x_{2}'\right)dx_{1}'dx_{2}'\end{split} & x_{1}>a_{1}\:\text{and}\:x_2>a_{2}
\\
\begin{split}
\stackrel[-\infty]{\infty}{\int}\stackrel[-\infty]{a_{1}}{\int}p\left(x_{1},x_{2},x_{1}',x_{2}'\right)dx_{1}'dx_{2}'\\
+\stackrel[-\infty]{a_{2}}{\int}\stackrel[a_{1}]{\infty}{\int}p\left(x_{1},x_{2},x_{1}',x_{2}'\right)dx_{1}'dx_{2}'
\end{split}& \text{else}
\end{cases}
\end{equation*}

\begin{figure}
\includegraphics[width=0.5\textwidth]{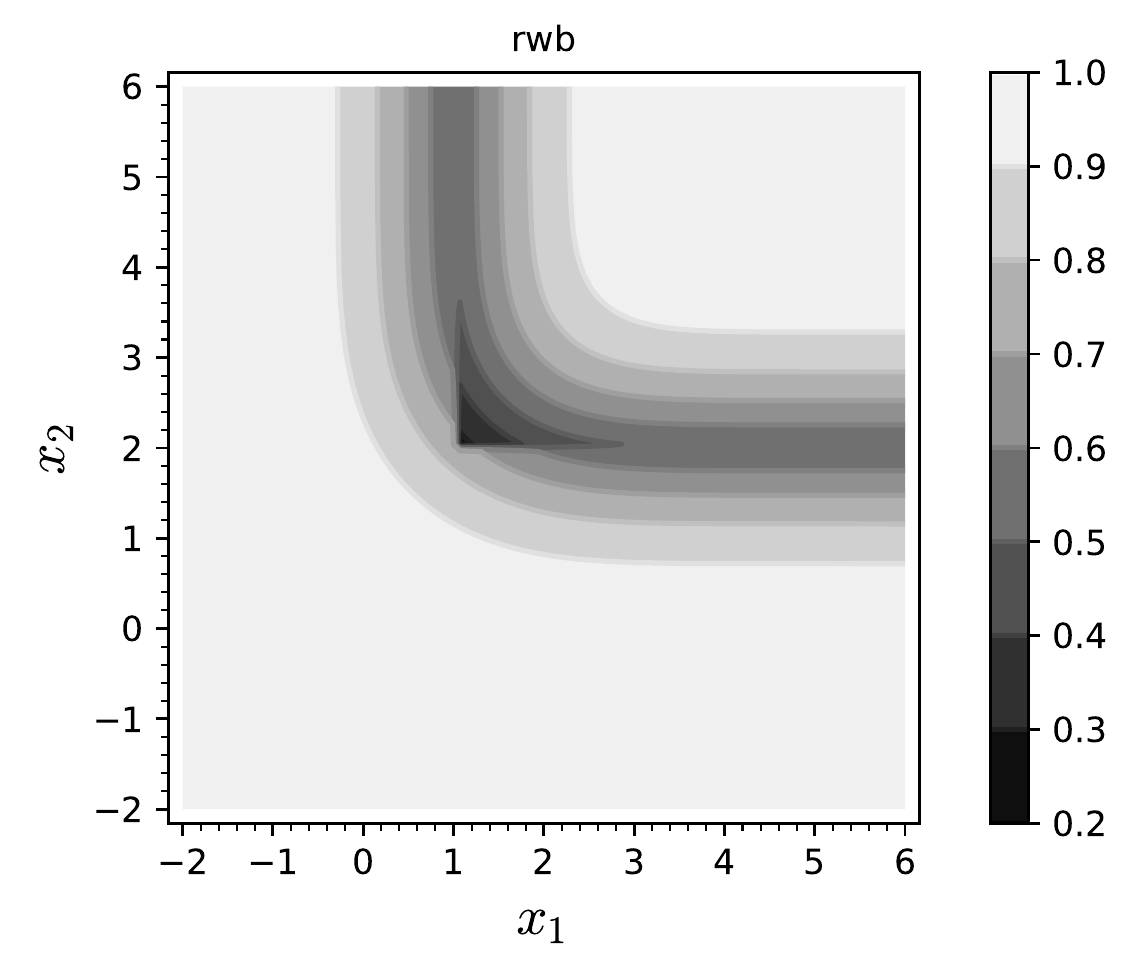}\includegraphics[width=0.5\textwidth]{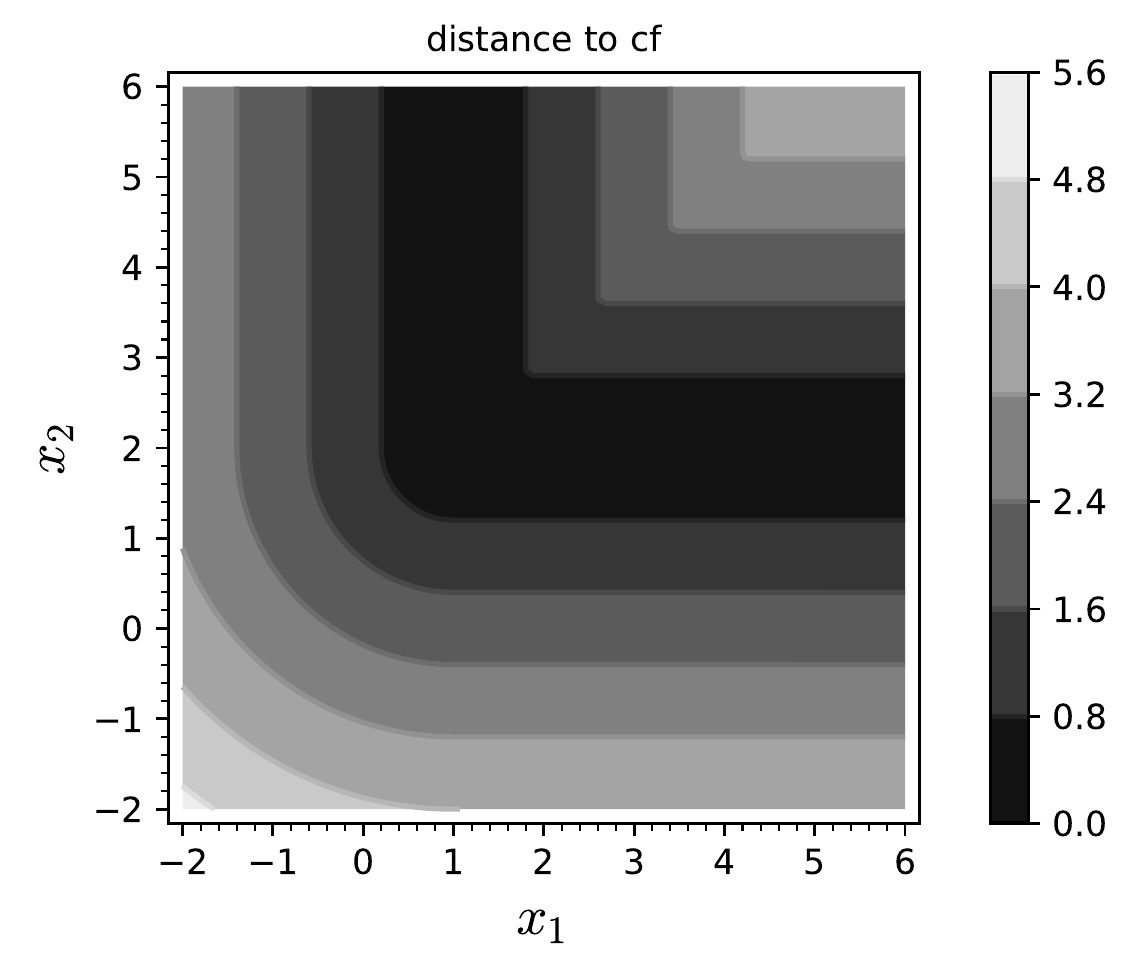}\caption{\label{fig:analytical-solution-corner}Analytical solution of real-world-robustness
(left) and distance to closest counterfactuel (right) for a 2-dimensional
linear decision boundary with $w_{1}=0,w_{2}=1,b=0$ and Gaussian
test-point uncertainty with $\sigma=1$.  Here the distance to the closest counterfactual is not normalized.}
\end{figure}

For uncertainty around the test points we use the same assumption as in the linear case (Eq. \ref{eq:gauss-uncertainy}). With this,
$P_{r}$ can be solved analytically again and yields

\begin{equation*}
P_{r}\left(x_{1},x_{2}\right)=\begin{cases}
\begin{split}
\frac{1}{4}\,{\left(\operatorname{erf}\left(\frac{\sqrt{2}x_{1}-\sqrt{2}}{2\,\sigma_{1}}\right)+1\right)}\operatorname{erf}\left(\frac{\sqrt{2}x_{2}-2\,\sqrt{2}}{2\,\sigma_{2}}\right)\\
+\frac{1}{4}\,\operatorname{erf}\left(\frac{\sqrt{2}x_{1}-\sqrt{2}}{2\,\sigma_{1}}\right)+\frac{1}{4}\end{split} & x_{1}>a_{1}\:\text{and}\:x_2>a_{2}\\
\begin{split}
-\frac{1}{4}\,{\left(\operatorname{erf}\left(\frac{\sqrt{2}x_{1}-\sqrt{2}}{2\,\sigma_{1}}\right)+1\right)}\operatorname{erf}\left(\frac{\sqrt{2}x_{2}-2\,\sqrt{2}}{2\,\sigma_{2}}\right)\\
-\frac{1}{4}\,\operatorname{erf}\left(\frac{\sqrt{2}x_{1}-\sqrt{2}}{2\,\sigma_{1}}\right)+\frac{3}{4} \end{split} & \text{else}
\end{cases}
\end{equation*}
Example plots for both robustness metrics are shown in fig. \ref{fig:analytical-solution-corner}.

Far away from the corner the information of both metrics is similar.
However, close the the corner, they diverge. The distance to the
closest counterfactual is only dependent on the smallest distance
(be it along $x_{1}$ or $x_{2}$). How much larger the distance along
the other axis is does not matter in this case. This is in contrast to our definition of real-world-robustness, where
it \emph{does} matter: In close vicinity to the edges the decision boundary is close in both directions, making the prediction less robust, as
it can change from perturbations along both axes. This is not correctly reflected when using the distance to the closest counterfactual as robustness metric.

\vskip 0.2in
\bibliography{real-world-robustness}

\end{document}